\pdfoutput=1  
\documentclass[11pt]{article}

\PassOptionsToPackage{numbers, sort&compress}{natbib}

\usepackage[utf8]{inputenc}
\usepackage[T1]{fontenc}
\usepackage[margin=1in]{geometry}
\usepackage{amsmath,amssymb,amsfonts,amsthm}
\usepackage{booktabs}
\usepackage{array}
\usepackage{tabularx}
\usepackage{multirow}
\usepackage{microtype}
\usepackage{graphicx}
\usepackage{xcolor}
\usepackage{paralist}
\usepackage{enumitem}
\usepackage{url}
\usepackage{hyperref}
\usepackage{nicefrac}
\usepackage{mathtools}
\usepackage{caption}
\usepackage{authblk}
\usepackage{natbib}

\hypersetup{colorlinks=true, linkcolor=blue, citecolor=blue, urlcolor=blue}

\theoremstyle{plain}

\theoremstyle{definition}

\theoremstyle{remark}

\newcolumntype{P}[1]{>{\centering\arraybackslash}p{#1}}

\newcommand{\R}{\mathbb{R}}

\newcommand{\norm}[1]{\left\lVert #1 \right\rVert}

\newcommand{\task}[1]{\texttt{#1}}
\newcommand{\prop}[1]{\textsf{#1}}

\newcommand{\vth}{\boldsymbol{\theta}}
\newcommand{\vg}{\mathbf{g}}
\newcommand{\vdel}{\boldsymbol{\delta}}
\newcommand{\vDel}{\boldsymbol{\Delta}}
\newcommand{\vh}{\mathbf{h}}
\newcommand{\vv}{\mathbf{v}}
\newcommand{\vr}{\mathbf{r}}
\newcommand{\vb}{\mathbf{b}}
\newcommand{\vq}{\mathbf{q}}

\usepackage{tikz}
\usetikzlibrary{positioning, arrows.meta, calc, fit, backgrounds, decorations.pathreplacing}
\definecolor{cblue}{rgb}{0.88,0.92,1.00}
\definecolor{camber}{rgb}{1.00,0.93,0.83}
\definecolor{cgreen}{rgb}{0.86,0.97,0.86}
\definecolor{cpurple}{rgb}{0.93,0.88,1.00}
\definecolor{dblue}{rgb}{0.25,0.40,0.75}
\definecolor{damber}{rgb}{0.80,0.55,0.15}
\definecolor{dgreen}{rgb}{0.25,0.55,0.30}
\definecolor{dpurple}{rgb}{0.45,0.35,0.70}
\definecolor{tsst}{HTML}{c23b22}
\definecolor{tboolq}{HTML}{2a6fbb}
\definecolor{trte}{HTML}{2e8b57}

\title{First-Order Predictable but Pairwise Fragile:\\Local Task Adaptation in Trained Transformers}

\author[1]{Irina Piontkovskaia}

\author[2,3]{Sergey Nikolenko}

\affil[1]{DAIMLD, Moscow, Russia}
\affil[2]{St. Petersburg Department of the Steklov Institute of Mathematics, St. Petersburg, Russia, \texttt{sergey@logic.pdmi.ras.ru}}
\affil[3]{St. Petersburg State University}

\date{\today}

\begin{document}

\maketitle

\begin{abstract}
Task arithmetic, sequential fine-tuning, activation steering, and first-order random search all operate through relatively small perturbations around an already trained checkpoint, and they rely on different local approximations: individual perturbations should be first-order predictable, task updates should compose with controlled interference, useful tangent structure should be stable and possible to estimate, and weight edits should have counterparts in representation space. We measure eight such properties with the same harness around the same multitask LoRA operating point, on nine transformers from seven families ($82$M--$7$B), with a prospectively registered property list, thresholds, and development/test split. We find a shared one-direction validity window up to the tested scale $10^{-2}$, but no universal radius for pairwise composition or update ordering. Along individual directions, changes of the probe loss (loss on a small fixed diagnostic batch) remain first-order predictable throughout the grid on every model median: a perturbation's effect on the loss is essentially its projection onto the gradient, which is also what makes local random search work. Pairwise structure, however, proves to be far more fragile: on over a third of the measured (model, task pair) combinations, two-update order sensitivity sets in strictly inside that window; the registered rank-$\le32$ mixture-covariance test turns out to be algebraically non-falsifiable at its sample count and shows no rank plateau up to $m=256$; task-gradient subspaces rotate within tens of steps; additivity under our fixed activation probe fails at full task-vector scale on several models, including both held-out $7$B models; and no model median passes the registered global mean-vector weight-to-steering correspondence bar. For two sequential task-gradient steps, the leading order-dependent term is the Lie bracket $H_B\vg_A-H_A\vg_B$. With the same minibatches on both sides, its normalized prediction $c(\eta)=\eta\kappa+O(\eta^2)$ tracks the measured defect at median ratio $1.002$, while the onset scale $\eta^\dagger\approx0.10/\kappa$ spans three orders of magnitude across models and task pairs. 
The bracket thus provides a precise, coordinate-specific consistency fact and a warning signal for update-order sensitivity; sample-stable prediction of task effects remains open.
\end{abstract}

\section{Introduction}
\label{sec:intro}

Many model-editing, merging, steering, and lightweight adaptation methods operate through relatively small perturbations around an already trained checkpoint, and each of them makes its own assumption about the local geometry there. Task arithmetic and model merging need functional additivity and controlled cross-task interference~\citep{ilharco2023editing,wortsman2022modelsoups,yadav2023tiesmerging}; sequential fine-tuning needs the order of updates not to matter too much; random parameter search needs useful first-order projections at the proposal scale~\citep{gan2026neural}; fixed task subspaces need tangent structure that is stable and estimable; activation steering needs weight-space changes to have predictable representation-space counterparts~\citep{turner2023actadd,panickssery2024caa,todd2024functionvectors,zou2023repe,wu2024reft}. These assumptions are similar, but they are far from identical: LoRA's low matrix rank~\citep{hu2022lora}, for instance, requires neither a low-rank stochastic-gradient covariance nor Euclidean flatness of its factor coordinates. Thus, natural questions arise: which approximation does each tool need, over what scale does it hold, and how should it be measured?

In a companion paper~\citep{piontkovskaia2026recoverable}, we asked which of these linear structures actually exist, and found a more subtle answer than we had expected: trained tasks can induce low-dimensional structure under specific per-task and trajectory-based estimators, but that structure is \emph{local and moving} rather than global and static. Static task planes miss the recovery direction, the useful basis drifts within a hundred steps, and a best-of-$N$ theorem that we have proved in~\citep{piontkovskaia2026recoverable} shows that isotropic random search picks up a dimension-independent amount of any useful direction, growing as $\sqrt{2\log N}$ in the budget, provided the loss is roughly linear in the perturbation. 

In~\citep{piontkovskaia2026recoverable}, the linear regime was found empirically at one scale on one model, which left all quantitative questions open: how linear is the regime, how low-dimensional, how additive, and to what scale does each of these hold across models? These questions can lead to practical conclusions: if you want to merge two adapters, compose two steering vectors, or perform a first-order search, you need to know where the relevant approximation holds and where it begins to fail. In this work, our goal is to answer these quantitative questions.

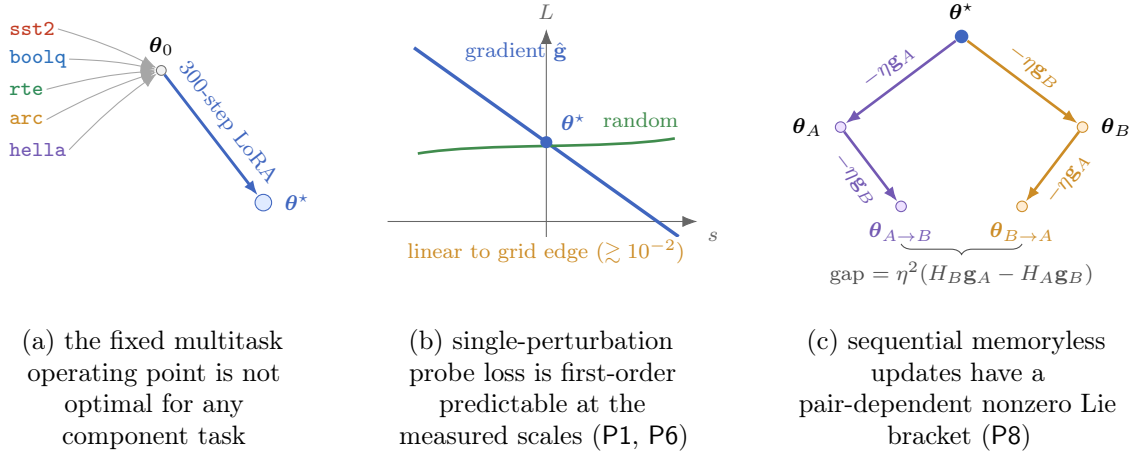
\begin{figure}[!t]
\centering
\setlength{\tabcolsep}{12pt}
\begin{tabular}{ccc}
\begin{tikzpicture}[font=\footnotesize, >={Latex[length=2mm]}, baseline=(current bounding box.center)]
  \useasboundingbox (-0.1,-0.2) rectangle (4.0,3.9);
  \foreach \y/\c/\n in {3.7/tsst/sst2, 3.3/tboolq/boolq, 2.9/trte/rte, 2.5/damber/arc, 2.1/dpurple/hella}{
    \node[right, text=\c, inner sep=1pt] (t\n) at (0.0,\y) {\scriptsize\texttt{\n}};
  }
  \node[circle, draw=black!55, fill=black!5, inner sep=1.3pt] (b) at (2.05,3.15) {};
  \node[above=0pt of b]{\scriptsize$\vth_0$};
  \node[circle, draw=dblue, fill=cblue, inner sep=2.2pt] (ts) at (3.4,1.4) {};
  \node[right=1pt of ts, text=dblue] {\scriptsize$\vth^\star$};
  \foreach \n in {sst2,boolq,rte,arc,hella}{ \draw[->, black!35, line width=0.4pt] (t\n.east) to[bend left=5] (b); }
  \draw[->, dblue, line width=1.1pt] (b) -- node[sloped, above, text=dblue]{\scriptsize $300$-step LoRA} (ts);
\end{tikzpicture}
&
\begin{tikzpicture}[font=\footnotesize, >={Latex[length=2mm]}, baseline=(current bounding box.center)]
  \useasboundingbox (-0.3,-0.2) rectangle (4.2,3.9);
  \draw[->, black!60] (-0.1,1.15) -- (4.0,1.15) node[below right]{\scriptsize $s$};
  \draw[->, black!60] (2.0,0.85) -- (2.0,3.7) node[above]{\scriptsize $L$};
  \draw[dblue, line width=1.3pt] (0.25,3.45) -- (3.75,0.95);
  \node[dblue] at (1.6,3.4) {\scriptsize gradient $\hat\vg$};
  \draw[dgreen, line width=1.0pt] (0.3,2.05) .. controls (1.2,2.2) and (2.8,2.1) .. (3.7,2.25);
  \node[dgreen] at (3.25,2.5) {\scriptsize random};
  \node[circle, fill=dblue, inner sep=1.6pt] (o) at (2.0,2.2) {};
  \node[above right=-1pt and 0pt of o, text=dblue]{\scriptsize$\vth^\star$};
  \draw[<->, damber, line width=1.0pt] (0.35,0.75) -- (3.65,0.75);
  \node[damber, fill=white, inner sep=1pt] at (2.0,0.75) {\scriptsize linear to grid edge ($\gtrsim10^{-2}$)};
\end{tikzpicture}
&
\begin{tikzpicture}[font=\footnotesize, >={Latex[length=2mm]}, baseline=(current bounding box.center)]
  \useasboundingbox (-0.3,-0.2) rectangle (4.5,3.9);
  \node[circle, fill=dblue, inner sep=1.8pt] (s) at (2.15,3.6) {};
  \node[above=0pt of s]{\scriptsize$\vth^\star$};
  \node[circle, draw=dpurple, fill=cpurple, inner sep=1.4pt] (a) at (0.55,2.4) {};
  \node[left=1pt of a]{\scriptsize$\vth_A$};
  \node[circle, draw=damber, fill=camber, inner sep=1.4pt] (b) at (3.75,2.4) {};
  \node[right=1pt of b]{\scriptsize$\vth_B$};
  \node[circle, draw=dpurple, fill=cpurple, inner sep=1.4pt] (ab) at (1.35,1.35) {};
  \node[circle, draw=damber, fill=camber, inner sep=1.4pt] (ba) at (2.95,1.35) {};
  \draw[->, dpurple, line width=1.0pt] (s) -- node[sloped,above]{\scriptsize$-\eta \vg_A$} (a);
  \draw[->, dpurple, line width=1.0pt] (a) -- node[sloped,below]{\scriptsize$-\eta \vg_B$} (ab);
  \draw[->, damber, line width=1.0pt] (s) -- node[sloped,above]{\scriptsize$-\eta \vg_B$} (b);
  \draw[->, damber, line width=1.0pt] (b) -- node[sloped,below]{\scriptsize$-\eta \vg_A$} (ba);
  \node[below=0pt of ab, text=dpurple]{\scriptsize$\vth_{A\to B}$};
  \node[below=0pt of ba, text=damber]{\scriptsize$\vth_{B\to A}$};
  \draw[decorate, decoration={brace, amplitude=4pt, mirror}, black!70] (ab.south) ++(0,-0.42) -- ($(ba.south)+(0,-0.42)$)
       node[midway, below=3pt]{\scriptsize gap $=\eta^2(H_B \vg_A - H_A \vg_B)$};
\end{tikzpicture}
\\[4pt]
\parbox[t]{4.0cm}{\centering\small (a) the fixed multitask operating point is not optimal for any component task} &
\parbox[t]{4.1cm}{\centering\small (b) single-perturbation probe loss is first-order predictable at the measured scales (\prop{P1}, \prop{P6})} &
\parbox[t]{4.4cm}{\centering\small (c) sequential memoryless updates have a pair-dependent nonzero Lie bracket (\prop{P8})}
\end{tabular}
\caption{Single-update predictability versus pairwise order dependence. All parameter-space statements use the fixed LoRA factor coordinates of the optimizer.}
\label{fig:setup}
\end{figure}

\begin{figure}[!t]
\centering
\includegraphics[width=\linewidth]{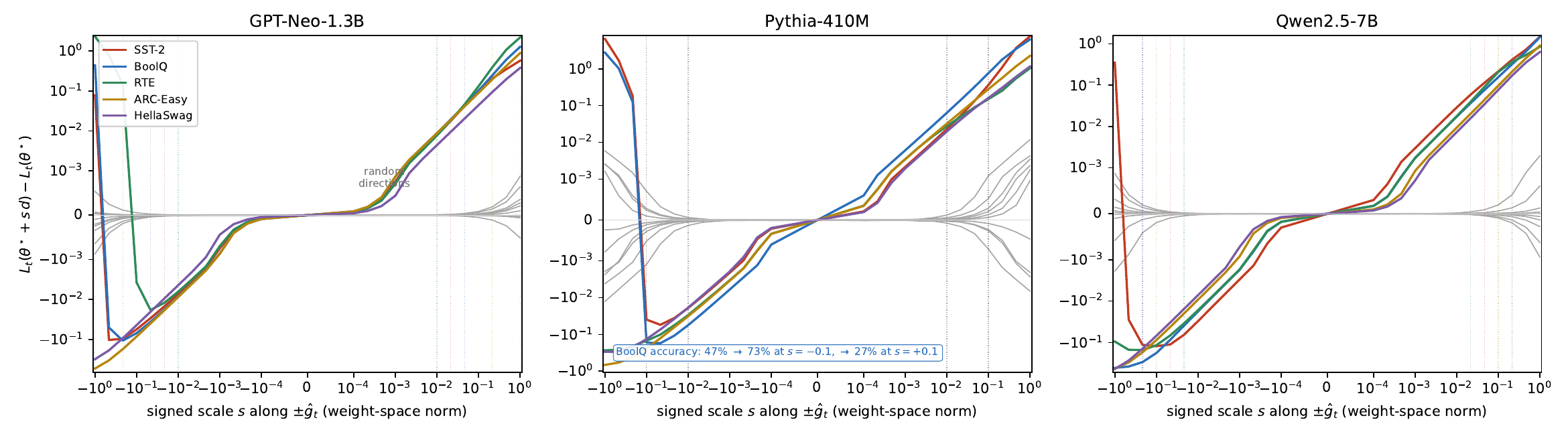}
\caption{Extended loss sweeps at $\vth^\star$ (one seed, three models): per-task probe loss along $\pm\hat\vg_t$ (colored), two fixed random unit directions (gray), and the largest contiguous tested segment with $A_t(s)<0.10$ (dotted verticals). Protocol and registered predictions in Section~\ref{sec:firstlook}.}
\label{fig:showcase}
\end{figure}

Specifically, we treat the local regime as a set of measurable properties and ask, for each: does it hold, to what scale does it hold, and is the answer the same across models? We consider eight properties (Section~\ref{sec:properties}): local linearity, low-dimensionality, tangent-space stability, additivity, parameter-to-activation correspondence, local searchability, directional curvature, and update non-commutativity. We execute them through the same frozen harness (the same code, estimator settings, and data streams for every model) around one multitask operating point, on nine models spanning seven architecture families and two orders of magnitude in size. Most of these quantities have been measured \emph{somewhere} in the literature, but on different models, tasks, and harnesses, so they cannot be compared against each other; measuring them together makes the evidence directly comparable. We have registered the property list, thresholds, and a development / test split before the sweep, reported every failure, and touched the three held-out models only once (see Appendix~\ref{app:frozen} for detailed provenance).

Before any statistical results, let us illustrate the object of study directly: Figure~\ref{fig:showcase} shows per-task loss profiles along each task's own gradient at $\vth^\star$. It is a one-direction view, and it gives a strongly anisotropic picture: along its own gradient, every task's loss is first-order predictable (the linear term of the Taylor expansion dominates the measured change) up to shifts of $10^{-2}$ and often far beyond, with a visible interior and a visible end, while along random directions the loss is flat until displacements an order of magnitude larger. Pairwise interactions between task updates are not shown in this figure; they are measured separately below (\prop{P4}, \prop{P8}), and most of the failures we find are pairwise.

One important empirical finding of our work is that two natural properties of this neighbourhood hold on very different scales. The first is \emph{single-update predictability}: along the tested directions, the change of the loss is well approximated by the first-order term $\vg^\top\vdel$ alone, and this approximation holds across the entire tested scale grid. 

The second concerns \emph{pairs} of task updates: when two updates are applied in sequence, the result can depend on their order. The interesting part is that this order dependence becomes substantial at a scale that is specific to the model and the task pair and that can fall \emph{inside} the range where single updates are still predictable. For two memoryless gradient steps, the leading order-dependent endpoint difference is the Lie bracket of the two tasks' gradient vector fields, and with our endpoint normalization the defect is
\[
c(\eta)=\eta\kappa+O(\eta^2),\qquad
\kappa=\frac{\norm{H_B\vg_A-H_A\vg_B}}{\norm{\vg_A+\vg_B}}.
\]

When both quantities are computed from the same frozen minibatches, the measured defect agrees with the leading term at median ratio $1.002$, so in fixed LoRA coordinates the onset of order dependence is set, for every model and task pair separately, by a cross-curvature that two Hessian-vector products can measure (Section~\ref{sec:spine}). This agreement verifies the second-order expansion at the step sizes we actually use, and that two independent routes to the same quantity (finite differences of real two-step updates and Hessian-vector products) give the same answer. 

But a stronger hypothesis fails: in a separately preregistered test (estimate the bracket on one data sample, build the two update orders from a second, evaluate task effects on a third), the bracket did not predict the held-out effects (Section~\ref{sec:crossfit}). Agreement on the same data therefore does not yield predictions across data. Thus, a shared first-order window makes local random search predictable, while pairwise task interactions often become fragile at much smaller, pair-specific scales.

We call the operating point \emph{primed}: it has been adapted to the task mixture, yet it still carries nonzero task-specific gradients, so every task leaves room for further local improvement. Controls that repeat the measurements at the raw pretrained checkpoint (Section~\ref{sec:controls}) find first-order predictability and the commutator mechanism there as well; priming shifts the measured scales, while the mechanisms are present from the start.

Table~\ref{tab:claims} collects the claims and their verdicts in one place, with pointers to where each is measured and qualified; we will refer back to it throughout the paper.

\begin{table}[!t]
\centering\small
\setlength{\tabcolsep}{4pt}
\caption{Claims at a glance: registered diagnostics and follow-up tests, their verdicts after all controls, and where each is measured and qualified.}
\label{tab:claims}
\begin{tabular}{p{0.40\linewidth}p{0.44\linewidth}l}
\toprule
Claim under test & Verdict & Where \\
\midrule
\prop{P1}: probe loss is first-order predictable along the gradient & holds on every model median (a shared window), right-censored at the grid edge $10^{-2}$ & \S\ref{sec:survivors} \\[1pt]
\prop{P6}: first-order scores rank random perturbations & holds and calibrated for $\sigma\le10^{-3}$ (in-sample); declines by $10^{-2}$ & \S\ref{sec:survivors} \\[1pt]
\prop{P7}: curvature along $\hat\vg$ exceeds a random direction & large ratio, confirmed against a $64$-direction distribution; sign not extremal & \S\ref{sec:survivors} \\[1pt]
\prop{P2}: mixture-gradient covariance is low-rank & no rank plateau up to $m=256$; the registered $m=32$ bar was non-falsifiable (sanity check only) & \S\ref{sec:failures} \\[1pt]
\prop{P3}: the task-gradient subspace is stable & fails: rotation within $16$--$80$ single-task steps & \S\ref{sec:failures} \\[1pt]
\prop{P4}: activation additivity at full task-vector scale & fails held-out ($1/3$); holds for $\alpha\lesssim0.3$ on all models; the failures persist under effective-weight addition & \S\ref{sec:failures} \\[1pt]
\prop{P5}: global mean-vector weight-to-CAA correspondence & fails on every model median & \S\ref{sec:failures} \\[1pt]
the three radii $\sigma^\star_1$, $\alpha^\star_4$, $\eta^\dagger$ coincide & registered comparison ill-posed; no shared scale on the common ruler & \S\ref{sec:ruler} \\[1pt]
\prop{P8}: the order defect follows $c(\eta)=\eta\kappa$ & holds with shared minibatches on both sides (median ratio $1.002$); onsets span three orders of magnitude & \S\ref{sec:matched} \\[1pt]
median threshold products carry to full FT and $13$--$14$B & hold at the median level; the pointwise mid-range check fails at full FT & \S\ref{sec:beyond} \\[1pt]
cross-fitted functional forecast from the bracket & not validated: aggregate invalid (step-size rule), eligible subset fails both bars & \S\ref{sec:crossfit} \\[1pt]
$\lambda_{\max}$ radius forecast on new models & not validated: undecidable on two of three models; inside/outside search comparison failed & \S\ref{sec:lammax} \\
\bottomrule
\end{tabular}
\end{table}

The remainder of the paper is organized as follows. Section~\ref{sec:related} reviews related work and reconciles the atlas with the companion paper. Section~\ref{sec:properties} fixes the operating point, the coordinate convention, and the eight diagnostics. Section~\ref{sec:firstlook} takes a direct look at the neighbourhood shown in Figure~\ref{fig:showcase}, Section~\ref{sec:atlas} reports the atlas across nine models and ends by putting every measured boundary on one displacement ruler. Section~\ref{sec:spine} studies the two-direction boundary: the Lie-bracket law, its verification, its extensions to full fine-tuning and $13$--$14$B scale, and the cross-fitted forecast (that fails). Section~\ref{sec:headroom} quantifies the first-order regime in terms of the loss headroom and search; Section~\ref{sec:practical} states what a practitioner can and cannot take from the atlas; Section~\ref{sec:discussion} discusses limitations, and Section~\ref{sec:conclusion} concludes the paper. Experimental details and pre-registration are described in the appendices.

\section{Related Work}
\label{sec:related}

\paragraph*{The companion paper.}
This work is the second paper of a line that treats small weight perturbations as displacements on the parameter manifold and studies the local structure that training installs~\citep{piontkovskaia2026recoverable}. The first paper established, on a controlled synthetic transformer and LoRA adapters with 1B-parameter experiments, that:
\begin{itemize}
\item per-task gradient structure is locally low-rank but the task plane drifts, so no static subspace captures adaptation; 
\item a best-of-$N$ theorem makes random search dimension-independent inside a locally linear window, which the work~\citep{piontkovskaia2026recoverable} located empirically (a scale near $10^{-4}$ on Qwen2.5-0.5B);
\item a weight move has an activation-space shadow which, in one gradient-step experiment, tracked a contrastive steering vector at cosine $0.58$.
\end{itemize}
In this work, we follows up on specific findings from the companion paper: we re-observe the tangent-plane motion (\prop{P3}), test the companion's best-of-$N$ formula at fixed radius, where the recoverable share is only $\sqrt{2\ln N/P}$, and study a two-update object not measured there. The companion's per-task covariance evidence remains supported in its original scope, whereas \prop{P2} that we consider here tests a mixture-gradient covariance at another operating point; likewise, \prop{P5} shows that one global, example-averaged SST-2 activation shadow does not reproduce the earlier single-cell cosine as a model-level rule, without ruling out input-conditioned or subspace correspondences.

\paragraph*{Local low-dimensional structure.}
\citet{li2018intrinsic} and \citet{aghajanyan2021intrinsic} show that training succeeds in random low-dimensional subspaces, bounding the intrinsic dimension of adaptation. Mode-connectivity work~\citep{garipov2018losssurface, draxler2018essentially} and loss-landscape studies~\citep{li2018visualizing, fort2019deep} expose further low-dimensional weight-space structure. These are distinct from \prop{P2}'s empirical rank of a mixture-gradient covariance: our control invalidates the original rank-$\le32$ conclusion for that estimator, and it does not adjudicate the other notions of low dimensionality.

\paragraph*{Editing, merging, and additivity.}
Task vectors~\citep{ilharco2023editing}, soups~\citep{wortsman2022modelsoups}, and TIES~\citep{yadav2023tiesmerging} motivate tests of composition, but simultaneous adapter composition and sequential optimization are different second-order objects, and we keep them separate: \prop{P4} measures activation additivity under one fixed probe, while \prop{P8} measures the antisymmetric order effect of two memoryless gradient steps. Recent PEFT work predicts pairwise merge retention from early alignment and representation signals~\citep{tang2026mergeability}, and partial linearization of adapter modules improves task-vector fusion~\citep{tang2023partial}; both reinforce the need for pair-specific diagnostics without identifying \prop{P4} with \prop{P8}.

\paragraph*{Sequential order and Lie brackets.}
The leading order effect $H_B\vg_A-H_A\vg_B$ is the Lie bracket of the two gradient fields, and it is not new to this paper. \citet{rukhovich2025commute} derive the bracket for domain-order interventions, discuss the noncanonical Euclidean parameter metric, and test HVP-based loss predictions in bilingual language-model pretraining. \citet{sweeney2026geometry} use the same primitive to predict transfer order in instruction tuning, preference optimization, pretraining domains, and longer schedules. Our contribution is a normalized endpoint defect, its finite-step calibration across LoRA operating points and task pairs, and its comparison with the other local-validity diagnostics measured at the same point. The scope is memoryless local gradient updates: optimizer state can introduce a leading $O(\eta)$ order term for fixed-clock momentum or AdamW~\citep{sweeney2026optimizer}.

\paragraph*{LoRA coordinates and merging geometry.}
A LoRA update has non-identifiable factors: $BA=(BQ)(Q^{-1}A)$ for invertible $Q$. Euclidean norms, random directions, gradients, Hessians, and curvature ratios in factor space can therefore change under a function-preserving gauge transformation. Symmetry-aware merging work treats low-rank adapters on a quotient manifold for exactly this reason~\citep{dasilva2026frechet}. Our measurements describe the fixed factor coordinates used by the optimizer; intrinsic function-space geometry is a separate object, which we leave open.

\paragraph*{Activation steering and the parameter--activation bridge.}
Activation addition~\citep{turner2023actadd}, contrastive activation addition~\citep{panickssery2024caa}, function vectors~\citep{todd2024functionvectors}, RepE~\citep{zou2023repe}, and ReFT~\citep{wu2024reft} steer in hidden-state space; ROME~\citep{meng2022rome} edits weights. The two are linked by the pushforward of a weight move into activation space, which our companion paper~\citep{piontkovskaia2026recoverable} stated as an identity and tested directly in one gradient-step setting~\citep{piontkovskaia2026recoverable}. Here, no model-level median passes the registered threshold for one specific global mean-vector construction. This cautions against treating weight updates and CAA mean vectors as interchangeable, but does not test an input-conditioned operator, an aligned subspace, or steering efficacy.

\paragraph*{Random search and zeroth-order methods.}
Evolution strategies~\citep{salimans2017evolution} and zeroth-order fine-tuning~\citep{malladi2023mezo} optimize without gradients; \citet{gan2026neural} use random parameter perturbation with top-$K$ selection and majority-vote ensembling as evidence of a dense thicket of task-improving specialists around pretrained weights, the phenomenon that motivated our line of work. Our own companion paper~\citep{piontkovskaia2026recoverable} supplied a Gaussian best-of-$N$ theorem; the present two-run check is consistent with its leading fixed-radius magnitude while making the $\sqrt{2\ln N/P}$ share of the gradient ceiling explicit.

\paragraph*{Hessian spectra and the bulk/outlier split.}
Empirical Hessian spectra of trained networks are often described as a low-rank outlier subspace, sometimes tied to class structure, plus a high-dimensional near-zero bulk~\citep{sagun2017hessian, papyan2019spectrum, papyan2019hierarchical, ghorbani2019investigation, gurari2018tiny}. \citet{song2025tiny} and \citet{dome2025} argue that useful signal can live away from the sharp outliers. As a stylized detectability model, a planted low-rank update against a random bulk has a BBP phase transition~\citep{bbp2005, paul2007, bgn2011, vershynin2010, montanari2026phase}. We use this empirical split as background motivation for where the two-direction boundary might come from but do not study it directly.

\paragraph*{Sharpness and curvature.}
Flatness and sharpness motivate many generalization and optimization studies~\citep{keskar2017largebatch, foret2021sam}, but sharpness is reparametrization-sensitive and not a standalone theorem of generalization~\citep{dinh2017sharp}. Our \prop{P7} statistic is more narrow: it compares signed curvature along the mixture-probe gradient with one random Rayleigh quotient in fixed LoRA coordinates.

\section{Setup and Diagnostics}
\label{sec:properties}

In this section we fix the operating point, the coordinate convention, and the eight diagnostics; everything later in the paper refers back to these definitions.

\paragraph{Operating point.}
We study adaptation around a single \emph{operating point} $\vth^\star$: a LoRA adapter~\citep{hu2022lora} (rank $16$, on the attention Q/K/V projections; exact per-family target modules in Appendix~\ref{app:details}) trained for $300$ steps of AdamW on a uniform mixture of five tasks (\task{sst2}, \task{boolq}, \task{rte}, \task{arc-easy}, \task{hellaswag}~\citep{boolq, arc, hellaswag}) in full precision with eager attention. All eight properties are measured at $\vth^\star$ (for the two that require a trajectory, on a single-task path leaving $\vth^\star$), so that every property describes the \emph{same} point of the \emph{same} trained network. The raw pretrained checkpoint (the \emph{base point}: the same network with the LoRA adapter at zero) is not the primary object, because the atlas is a map of adaptation \emph{around a trained point}: task vectors, local search directions, and update order are all read from a point that has already absorbed the multitask mixture. 

Note that, unlike the companion paper~\citep{piontkovskaia2026recoverable}, there is no forgetting/recovery recipe here: $\vth^\star$ is just ``a trained network'', which keeps the setup uniform across models that do and do not forget under such recipes. Section~\ref{sec:controls} reports a control in which the battery is repeated at the raw base checkpoint; it measures what the $300$ AdamW steps change without claiming a unique temporal regime.

We write $\vth\in\R^P$ for the trainable (LoRA) parameters, $L(\vth)$ for the mean next-token loss on a fixed probe set, $\vg=\nabla L(\vth^\star)$, and $H=\nabla^2 L(\vth^\star)$. The same fixed multitask probe set (four frozen batches of eight examples each, $32$ examples total, sampled task-uniformly from the training pools; Appendix~\ref{app:details}) defines $L$, $\vg$, and $H$, so that first- and second-order checks are mutually consistent; the one-direction properties use the mixture gradient for exactly this reason (Section~\ref{sec:controls} reports the per-task-probe control). Null comparators differ by property: \prop{P2} uses an isotropic reference, \prop{P3} a same-checkpoint resampling floor, \prop{P4}/\prop{P5}/\prop{P7} norm-matched random controls, and \prop{P6} a first-order prediction against realized Gaussian draws; \prop{P1} and \prop{P8} are threshold curves.

\paragraph{Coordinate convention.}
There is one important convention that concerns every claim about the parameter space in the paper. All norms, isotropic draws, gradients, Hessian eigenvalues, directional curvatures, $\kappa$ values, and displacement radii are Euclidean quantities in the fixed coordinates of the trained LoRA factors. If an effective update is $\Delta W=BA$, then $(B,A)$ and $(BQ,Q^{-1}A)$ represent the same function for every invertible $Q$ but need not have the same Euclidean geometry. Our results therefore describe the geometry as seen by the stated optimizer, not an intrinsic geometry of functions or effective weight updates. In particular, \prop{P4} adds factor-coordinate displacements rather than effective $\Delta W$ matrices.

\paragraph{Three distinct second-order objects.}
The diagnostics separate self-curvature from symmetric and antisymmetric cross-interactions; let us set up this division explicitly because it explains why the measured scales need not coincide. For a unit gradient direction, Taylor expansion gives
\[
A(\sigma)=\frac{\sigma\,|\hat\vg^\top H\hat\vg|}{2\norm{\vg}}+O(\sigma^2),
\]
so \prop{P1} and the numerator of \prop{P7} deal with the same \emph{self-curvature} through different routes.

For a twice differentiable activation map, \prop{P4}'s numerator begins with the \emph{symmetric} mixed derivative $\alpha^2D^2\vh[\vdel_A,\vdel_B]$. In contrast, \prop{P8}'s order-sensitive endpoint begins with the \emph{antisymmetric} Lie bracket $\eta^2(H_B\vg_A-H_A\vg_B)$. These are contractions of different tensors, evaluated through different observables, and they do not have to have the same scale. Informally speaking, a move along one direction feels its own curvature, two moves applied together feel the symmetric part of their interaction, and two moves applied one after the other feel the antisymmetric part. In this work, we try to keep these three questions separate, and the measurements below show that their scales are indeed different. Moreover, since \prop{P4} adds LoRA factor displacements, its effective weight contains bilinear cross terms $\delta B_A\delta A_B+\delta B_B\delta A_A$ that are absent when dense deltas of effective weights are added; a follow-up control reported with the \prop{P4} results (Section~\ref{sec:failures}) separates this coordinate effect from network nonlinearity.

\paragraph{The eight diagnostics.}
Each property is a statistic with a frozen threshold; we state them here in compact form, with estimator details in Appendix~\ref{app:details}.

\begin{itemize}[leftmargin=1.6em]
\item \prop{P1} (\emph{local linearity}). Along the unit gradient direction $\hat\vg$, the \emph{antisymmetry ratio} at scale $\sigma$ is
\[
A(\sigma)=\frac{\bigl|L(\vth^\star+\sigma\hat\vg)+L(\vth^\star-\sigma\hat\vg)-2L(\vth^\star)\bigr|}{\bigl|L(\vth^\star+\sigma\hat\vg)-L(\vth^\star-\sigma\hat\vg)\bigr|},
\]
the ratio of the symmetric (curvature) part to the antisymmetric (linear) part of the loss change; this is the same decomposition the companion paper~\citep{piontkovskaia2026recoverable} used to locate the linear regime. The frozen primary statistic is $\sigma^\star_1=\max\{\sigma: A(\sigma)<0.10\}$, the largest passing grid point. This is a lower-bound diagnostic: at the smallest scales the denominator can hit the numerical floor, and some seed curves fail there before passing at larger scales, so $\sigma^\star_1$ should not be read as a contiguous-interval radius. The extended sweeps of Figure~\ref{fig:showcase} support reading these small-scale irregularities as a noise floor: the profiles there are visually smooth and approximately antisymmetric on a contiguous central segment around zero.
\item \prop{P2} (\emph{low-dimensionality}). From $m$ per-minibatch mixture gradients at $\vth^\star$ we form the centered empirical covariance and report $r_{90}$, the number of components carrying $90\%$ of its energy. The substantive analysis is a sample-count sweep over $m\in\{32,64,128,256\}$ asking whether $r_{90}$ stabilizes as $m$ grows (Section~\ref{sec:failures}). 
\item \prop{P3} (\emph{tangent-space stability}). Along a single-task trajectory, let $U_t$ be the estimated top-$10$ task-gradient subspace and let $\rho(U,V)=\norm{U^\top V}_F^2/10$ be the normalized overlap (larger = more stable). At each checkpoint we compare $\rho(U_8,U_t)$ with the same-checkpoint resampling overlap $\rho(U_t,U'_t)$; the rotation onset is the first checkpoint where the former falls at least $0.02$ below the latter.
\item \prop{P4} (\emph{additivity}). For task-vector pairs $(\vdel_A,\vdel_B)$ (short single-task fine-tunes from $\vth^\star$) and fraction $\alpha$, the \emph{activation-space composition error} is
\[
\varepsilon_{\rm add}(\alpha)=\frac{\norm{\vh(\alpha(\vdel_A+\vdel_B))-\vh(\alpha\vdel_A)-\vh(\alpha\vdel_B)}}{\norm{\vh(\alpha\vdel_A)}+\norm{\vh(\alpha\vdel_B)}},
\]
where $\vh(\vdel)$ is the change in the upper-medium-layer last-token activation induced by applying $\vdel$. \prop{P4} is a systematic test across models and task pairs, but under one fixed activation probe: the observable is the mean last-token hidden state at one layer on $16$ fixed SST-2 prompts, with one task-vector recipe and three registered pairs. It does not test general function additivity, logits, accuracy, or effective-weight addition. We report $\alpha^\star_4$, the largest passing $\alpha$ on the grid under $\varepsilon_{\rm add}<0.15$, with $\alpha^\star_4=1.0$ a max-grid lower bound.
\item \prop{P5} (\emph{parameter--activation correspondence}). For a single gradient step $\vdel\vth=-\eta\vg$ at $\eta=10^{-4}$, the \emph{activation shift} is the mean change in the upper-medium-layer last-token hidden state; we report its cosine to the labelled-contrast (CAA) steering vector~\citep{panickssery2024caa} built at the same layer, taking the best of three candidate layers. The correspondence holds if this cosine exceeds $0.30$.
\item \prop{P6} (\emph{local searchability}). For isotropic Gaussian perturbations $\vdel\sim\mathcal N(0,\sigma^2 I)$ at per-coordinate scale $\sigma=10^{-3}$ (total norm $\sigma\sqrt{P}$, of order $1$ for these adapters; a different axis from \prop{P1}'s unit-direction scale), we report the correlation between the realized gain $L(\vth^\star)-L(\vth^\star+\vdel)$ and the first-order prediction $-\vg^\top\vdel$, plus the fraction of beneficial draws. The regime is \emph{searchable} if the correlation exceeds $0.5$.
\item \prop{P7} (\emph{directional curvature}). Via autograd Hessian-vector products, we compare signed curvature $\hat\vg^\top H\hat\vg$ along the mixture-probe gradient with $\vr^\top H\vr$ along one random unit vector, subject to a finite-difference HVP reconstruction check; the frozen statistic is $|\hat\vg^\top H\hat\vg/\vr^\top H\vr|>3$. A single random Rayleigh quotient can be near zero through spectral cancellation, so the ratio indicates anisotropy relative to that one comparator; it does not estimate absolute sharpness.
\item \prop{P8} (\emph{non-commutativity}). For one update of size $\eta$ on task $A$ then $B$ versus $B$ then $A$, the \emph{commutator defect} is
\[
c(\eta)=\frac{\norm{\vDel_{A\to B}(\eta)-\vDel_{B\to A}(\eta)}}{\norm{\vDel_{A\to B}(\eta)}},
\]
and we report the onset $\eta^\dagger=\min\{\eta: c(\eta)\ge 0.10\}$. \prop{P8} is a weight-space path property (do two sequential optimizations reach the same endpoint?), while \prop{P4} is an activation-space composition property; $\eta$ is a per-task learning rate, and the induced parameter step has norm $\approx\eta\norm{\vg_A+\vg_B}$, again a different axis from \prop{P1}'s $\sigma$.
\end{itemize}

\paragraph{Protocol constants.}
LoRA rank $16$, $\alpha=32$, dropout $0$, on the attention Q/K/V projections; $300$ AdamW steps at $2\cdot10^{-4}$, training batch $8$, sequence length $\le160$. The probe loss (defining $L$, $\vg$, $H$) is a fixed set of four eight-example multitask batches ($32$ examples). \prop{P2} uses $m=32$ per-minibatch gradients in the main battery, with a follow-up $m$-sweep at $m\in\{32,64,128,256\}$; \prop{P6} uses $256$ Gaussian draws; \prop{P3} tracks $200$ single-task steps with subspaces re-estimated at checkpoints $\{8,16,32,48,64,96,128,160,200\}$; \prop{P4} task vectors are $40$-step single-task fine-tunes swept over $\alpha\in\{0.03,0.1,0.3,1.0\}$; \prop{P1} and \prop{P8} use the scale grid $\{10^{-5},3\cdot10^{-5},10^{-4},3\cdot10^{-4},10^{-3},3\cdot10^{-3},10^{-2}\}$ on their own axes; the task pairs for \prop{P4}/\prop{P8} are (\task{sst2},\task{boolq}), (\task{sst2},\task{rte}), (\task{boolq},\task{rte}). The battery uses three seeds per (model, property) pair. Throughout the paper we call one point of this measurement grid---either one (model, seed, task-pair) combination or a (model, seed) where no pair is involved---a \emph{cell}. Full details, including the model list, prompt templates, probe construction, per-property estimator settings, layer choices, and hardware, are collected in Appendix~\ref{app:details}.

\paragraph{Registration.}
This is a part confirmatory, part exploratory project. The property list, model split, keep/drop rule, and threshold bars were prospective; estimator details were frozen in code; the fine-grid spine studies and the controls were registered separately before their data was read; the headroom and common-ruler analyses are descriptive reanalyses of archived measurement records. We report all properties, including the ones that fail, and we touched the three held-out atlas models once, after the keep/drop decisions were frozen. Appendix~\ref{app:frozen} gives the provenance analysis by analysis, and Appendix~\ref{app:thresholds} shows that the headline patterns are stable under nearby threshold choices.

\section{A First Look at the Neighbourhood}
\label{sec:firstlook}

Before systematic measurements, let us look directly at the neighbourhood they will quantify. Figure~\ref{fig:showcase} plots, for three models (two development, one held-out family) and all five tasks at $\vth^\star$, the per-task probe loss along $\pm$ the task's own unit gradient over an extended grid $s\in[10^{-4},1]$, reaching well past the atlas ceiling of $10^{-2}$, so that both the start and the end of the linear window are visible. It also shows the loss along two fixed random unit directions, and multiple-choice accuracy probed at selected points; the dotted verticals mark, per task, the largest contiguous tested segment with antisymmetry ratio $A_t(s)<0.10$. The panels are easiest to read from the center outwards: both axes are signed log scales, so a first-order response appears as two roughly antisymmetric wings around $s=0$, curvature shows up as the wings losing that symmetry, and the flat gray band through the middle is what ``random directions do nothing'' looks like at these scales. The protocol and its three predictions were pre-registered before the run; all three scored as hits.

First, on every model the along-gradient loss is linear ($A_t<0.10$) through $s=10^{-2}$ on all five tasks ($15/15$ cells), and it departs from linearity by $s=0.3$ on $11/15$: the window has a visible interior and a visible end, with edges ranging from $s=0.01$ to beyond the grid (HellaSwag on GPT-Neo is still linear at $s=1$). Second, at $s=10^{-2}$ the along-gradient loss change exceeds the change along each of two fixed random directions by $480\times$--$4300\times$ per cell; the random curves are visually flat until $|s|\gtrsim0.1$. (A one-seed directional comparison; it does not by itself validate \prop{P7}'s single-denominator ratio.) Third, accuracy stays within $\pm2$pp at the $s=10^{-2}$ window edge on $13/15$ cells, as registered; the largest response is BoolQ on Pythia-410M, which moves $+10$pp at downhill norm $0.01$ and runs from $73\%$ downhill to $27\%$ uphill at $|s|=0.1$, a $46$pp spread between the two directions, while RTE on GPT-Neo gains $+3.3$pp at the edge. Accuracy uses the first $55$--$64$ usable validation examples per task, so a single example is worth $1.6$--$1.8$pp, and shifts of $\pm2$--$3$pp (including the RTE one, which is two examples) are within counting noise; the BoolQ response is far outside it. Appendix~\ref{app:showcaseacc} tabulates the per-cell accuracy results at every probed scale, in both directions.

We can draw three lessons from this figure. The one-direction window is wide and its edges are heterogeneous: within a single model they vary by two orders of magnitude across tasks. Loss sensitivity and accuracy sensitivity can be very different: on most cells the loss slope moves accuracy by almost nothing at the window edge, while a single cell converts the same displacement into tens of points. Finally, every curve here is one direction at a time; failures that matter for task composition appear only when two updates are combined.

\section{The Atlas Across Nine Models}
\label{sec:atlas}

We run the battery on six development models spanning five families: DistilGPT-2~\citep{huggingface2019distilgpt2}, Pythia-160M~\citep{biderman2023pythia}, Pythia-410M, GPT-Neo-1.3B~\citep{black2021gptneo}, OPT-1.3B~\citep{zhang2022opt}, and TinyLlama-1.1B~\citep{zhang2024tinyllama}. We make all keep/drop decisions there, freeze them, and then run the same battery once on three held-out models: Pythia-1.4B (scale within a seen family), Qwen2.5-7B~\citep{qwen2024qwen25} (an unseen family at scale), and OLMo-7B-Instruct~\citep{groeneveld2024olmo} (an unseen family on which activation steering transported poorly in the companion paper's experiments). Table~\ref{tab:atlas} shows the per-model medians for all eight diagnostics on all nine models; a star there records only that the originally registered numeric bar was passed, and the discussion below states what survived the controls. We discuss failures first.

\begin{table}[!t]
\centering\footnotesize
\setlength{\tabcolsep}{1.5pt}
\caption{Atlas measurements: per-model medians over seeds. A star records passage of the originally registered numeric bar only; $\mathrm{nf}$ = non-falsifiable (\prop{P2}, superseded by the $m$-sweep), $\dagger$ = right-censored at the grid maximum, $\ddagger$ = coarse-grid onset refined later, $^{\rm in}$ = failed HVP reconstruction check. Per-seed values in Appendix~\ref{app:seeds}.}
\label{tab:atlas}
\begin{tabular}{llcccccc|ccc}
\toprule
 & & \multicolumn{6}{c|}{\textbf{Development}} & \multicolumn{3}{c}{\textbf{Held-out}} \\
 & \textbf{Diagnostic} & distil & pyt-160m & pyt-410m & neo-1.3B & opt-1.3b & TinyLl-1.1B & pyt-1.4B & Qwen-7B & OLMo-7B \\
\midrule
\prop{P1} & linearity $\sigma^\star_1$ & $10^{-2\dagger}$ & $10^{-2\dagger}$ & $10^{-2\dagger}$ & $10^{-2\dagger}$ & $10^{-2\dagger}$ & $10^{-2\dagger}$ & $10^{-2\dagger}$ & $10^{-2\dagger}$ & $10^{-2\dagger}$ \\
\prop{P2} & low-dim $r_{90}$ ($m{=}32$) & $23^{\rm nf}$ & $24^{\rm nf}$ & $24^{\rm nf}$ & $21^{\rm nf}$ & $23^{\rm nf}$ & $23^{\rm nf}$ & $23^{\rm nf}$ & $23^{\rm nf}$ & $21^{\rm nf}$ \\
\prop{P3} & rotation onset (steps) & $80$ & $16$ & $16$ & $32$ & $16$ & $16$ & $16$ & $16$ & $16$ \\
\prop{P4} & additivity $\varepsilon(\alpha{=}1)$ & $0.07\ast$ & $0.14\ast$ & $0.08\ast$ & $0.19$ & $0.35$ & $0.14\ast$ & $0.11\ast$ & $0.32$ & $0.32$ \\
\prop{P5} & pushforward $\cos$ & $+.01$ & $+.01$ & $-.12$ & $+.05$ & $+.22$ & $+.04$ & $-.22$ & $+.01$ & $-.06$ \\
\prop{P6} & searchability corr & $.77\ast$ & $.74\ast$ & $.97\ast$ & $1.0\ast$ & $.70\ast$ & $1.0\ast$ & $1.0\ast$ & $1.0\ast$ & $.97\ast$ \\
\prop{P7} & curvature $|$ratio$|$ & $1.7e3^{\rm in}$ & $1.7e5\ast$ & $3.1e4\ast$ & $1.9e4\ast$ & $7.0e4^{\rm in}$ & $2.1e5\ast$ & $1.7e4\ast$ & $4.0e4\ast$ & $2.4e5\ast$ \\
\prop{P8} & commutator $\eta^\dagger$ & $10^{-5}$ & $3{\cdot}10^{-5}$ & $3{\cdot}10^{-3}$ & $10^{-2\,\ddagger}$ & $10^{-5}$ & $10^{-2}$ & $10^{-2\,\ddagger}$ & $10^{-2\,\ddagger}$ & $3{\cdot}10^{-3}$ \\
\bottomrule
\end{tabular}
\end{table}

\subsection{What fails}
\label{sec:failures}

\begin{figure}[!t]
\centering
\includegraphics[width=0.6\linewidth]{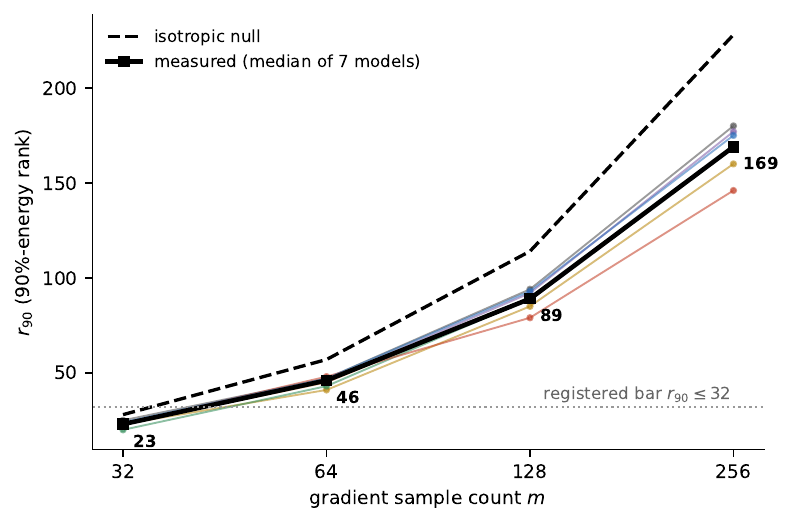}
\caption{\prop{P2} sample-count control: $r_{90}$ against $m$ for the seven models (thin lines, seed 0), their median (bold), and the matched isotropic null (dashed).}
\label{fig:msweep}
\end{figure}

\paragraph{The mixture-gradient covariance shows no low-rank plateau.}
The substantive \prop{P2} question is whether the mixture-gradient covariance concentrates in a small subspace whose dimension stops growing once enough samples are seen. It does not: in a seed-0, seven-model sweep, $r_{90}$ grows $23\to46\to89\to169$ as $m$ grows $32\to64\to128\to256$, with no plateau in range, while remaining more concentrated than a matched isotropic null (ratio $0.82\to0.74$; Figure~\ref{fig:msweep}). The originally registered bar, $r_{90}\le32$ at $m=32$, was satisfied automatically and carries no evidential weight: thirty-two samples can span at most thirty-one directions after centering, so asking whether they concentrate in thirty-two was never a question about the network, and the apparent universal value $r_{90}\approx21$--$25$ in Table~\ref{tab:atlas} is a sanity check of the implementation. The safe conclusion is estimator-specific: the apparent rank-$\le32$ mixture-gradient structure was a sample-count artifact. This does not negate low matrix rank of LoRA updates, intrinsic optimization dimension, trajectory dimension, or the companion paper's per-task covariance result (a mixture covariance can be high-dimensional even when every conditional task covariance is concentrated), and the within-task/between-task decomposition remains to be measured. A separately preregistered task-conditioned follow-up stopped at a failed whole-matrix numerical audit, before any scientific scoring; it yields no verdict, and we use none of its partial metrics.

\begin{figure}[!t]
\centering
\includegraphics[width=\linewidth]{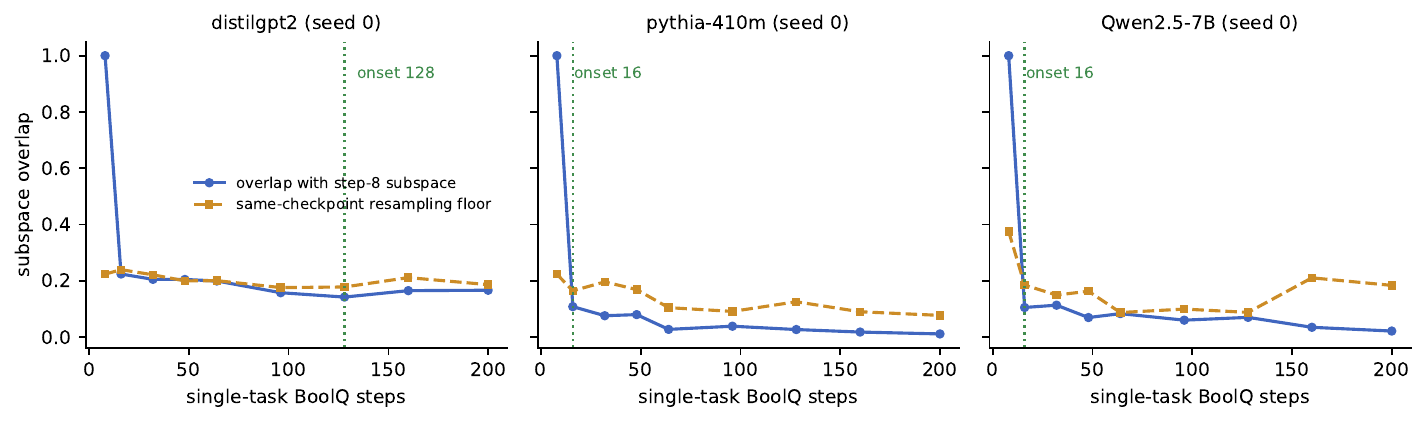}
\caption{\prop{P3} on the BoolQ trajectory (seed 0): overlap with the step-8 subspace against the same-checkpoint resampling floor.}
\label{fig:radii}
\end{figure}

\paragraph{The tangent space rotates within tens of steps.}
\prop{P3} fails in its original ``stability'' form, and we expected that when registering it: the main result of the companion paper was precisely that the task plane moves. On the BoolQ trajectory where the diagnostic is available, model medians clear the same-checkpoint resampling floor within $16$--$80$ single-task steps (Figure~\ref{fig:radii}), after which a small core persists (asymptotic overlap $0.02$--$0.17$). DistilGPT-2, the leftmost panel, is the slowest rotator in the battery, and even it reaches the floor by step $\sim$$100$; note also that the floor itself sits well below $1$, so part of the lost overlap is estimator variance, which is why the onset criterion compares drift to the floor. The exact timing depends on the seed (DistilGPT-2's three seeds give $128/\text{---}/32$; Appendix~\ref{app:seeds}), so the supported claim is early rotation on this trajectory, not a universal clock. In practice, a method that assumes a fixed task subspace should recheck that assumption within tens of steps.

\paragraph{The registered global mean-vector correspondence does not generalize.}
For \prop{P5}, model-median cosines between one SST-2 gradient step's mean activation shadow and one labelled-contrast CAA mean vector range from $-0.22$ to $+0.22$: no model median reaches the $0.30$ bar, including all three held-out models. The companion paper's $0.58$ cell used another model, operating point, and prompt construction. The rejected hypothesis is also quite specific: a \emph{global, example-averaged} activation shadow aligning with a \emph{global} CAA mean vector. Individual seeds do exceed the bar (up to $+0.34$), and input-conditioned maps, aligned subspaces, and steering efficacy remain untested here. The pushforward $x\mapsto D_{\vth} \vh(x;\vth)\,\vdel\vth$ is input-dependent, and averaging it into one vector can erase structure through cancellation; the cheapest version of the bridge does not survive averaging, while the companion's input-level identity is untouched.

\begin{figure}[!t]
\centering
\includegraphics[width=0.78\linewidth]{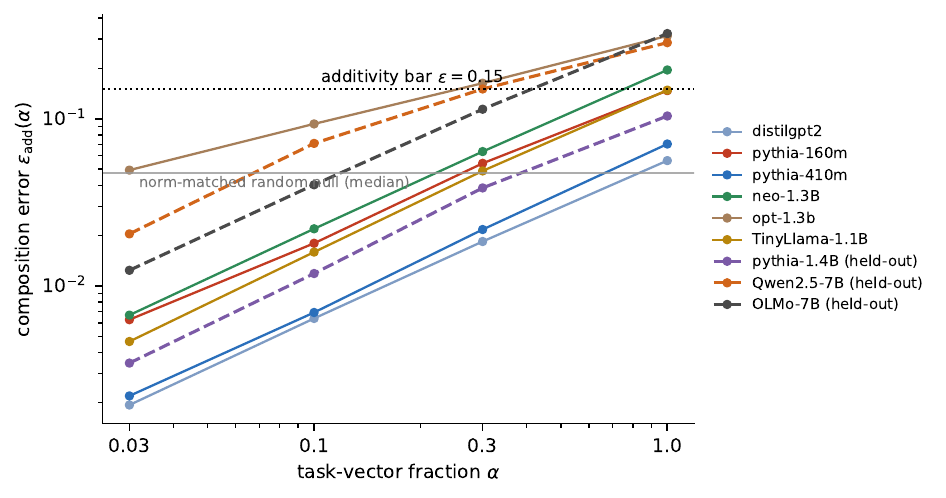}
\caption{\prop{P4} composition error $\varepsilon_{\rm add}(\alpha)$, per-model median over pairs and seeds (log--log); held-out models dashed, norm-matched random null in gray.}
\label{fig:p4}
\end{figure}

\paragraph{Activation additivity fails at full task-vector scale on several models.}
Under \prop{P4}'s fixed SST-2, one-layer activation probe, four of six development models pass at $\alpha=1$, but the registered held-out bar fails ($1/3$): both held-out $7$B models have composition error $0.32$, and GPT-Neo-1.3B and OPT-1.3B also fail the full-scale bar. Figure~\ref{fig:p4} shows the whole picture, and it is more regular than the pass/fail column suggests: on every model the error grows essentially linearly in $\alpha$, as the symmetric mixed-derivative term predicts, so the models differ mainly in the \emph{coefficient}, and ``passing at $\alpha=1$'' just means having a coefficient below the bar. Every model passes for $\alpha\lesssim0.3$. Under this probe, additivity at full scale is not model-universal. A follow-up control separates the coordinate effect from network nonlinearity: it materializes each task's effective-weight delta $\Delta W$ and adds the matrices, so the factor cross terms vanish (the two routes agree on single tasks to relative error below $10^{-2}$ on all $81$ cells). The coordinate choice accounts for part of the measured error, with the median $\varepsilon(1)$ dropping from $0.180$ to $0.131$, but not for the verdicts: of the five models whose factor-route median reaches the $0.15$ bar in this rerun, only TinyLlama-1.1B falls below it under effective-weight addition. The full-scale additivity failures are mostly network nonlinearity, further inflated by the factor coordinates.

\subsection{What survives}
\label{sec:survivors}

Two first-order diagnostics pass their frozen bars on every model median, and one second-order diagnostic survives with qualifications.

\paragraph{The loss is first-order predictable along the gradient.}
\prop{P1}'s antisymmetry ratio stays below $0.10$ out to the largest tested scale ($\sigma=10^{-2}$) on every model median, development and held-out alike. All nine medians are therefore \emph{right-censored} at the grid edge and should be read as lower bounds; the extended sweeps of Section~\ref{sec:firstlook} show that the window does end farther out, between $s=0.01$ and past $s=0.3$ depending on the cell. The seed-level picture is rougher: one Pythia-160M seed has $\sigma^\star_1=10^{-4}$, and eight of $27$ development seed curves are nonmonotone at the smallest scales, where the denominator approaches the numerical floor. The supported statement is therefore that the model-median mixture-probe curves remain first-order dominated at $10^{-2}$, not that every seed is linear on a contiguous interval.

\paragraph{Random-perturbation gains are first-order predictable at the tested scale.}
\prop{P6}'s correlation between realized probe gain and $-\vg^\top\vdel$ is $0.70$--$1.00$ across the nine model medians. This is the practical payoff of \prop{P1}: within the window, a perturbation's effect is essentially its projection onto the gradient, so the sign and size of a random proposal's gain can be read off in advance, roughly half of isotropic draws carry a downhill component, and best-of-$N$ selects the largest projection. Because the same $32$ examples define $\vg$ and evaluate the gain, this is an in-sample Taylor-ranking check at one proposal scale: it shows that the probe loss is locally linear enough to rank random proposals. A follow-up sweep over the full scale grid maps the range: the correlation stays above the $0.5$ bar at every $\sigma\le10^{-3}$ on all $23$ measured cells, the calibration slope of gain against prediction lies in $[0.5,1.5]$ on every cell for $\sigma\le3\cdot10^{-4}$, and the correlation declines from $10^{-3}$ to $10^{-2}$ on all nine models (on OLMo-7B from $0.99$ to $0.46$). Generalization of score-selected perturbations to disjoint data remains untested; Section~\ref{sec:headroom} adds a two-model descriptive magnitude check.

\begin{figure}[!t]
\centering
\includegraphics[width=0.72\linewidth]{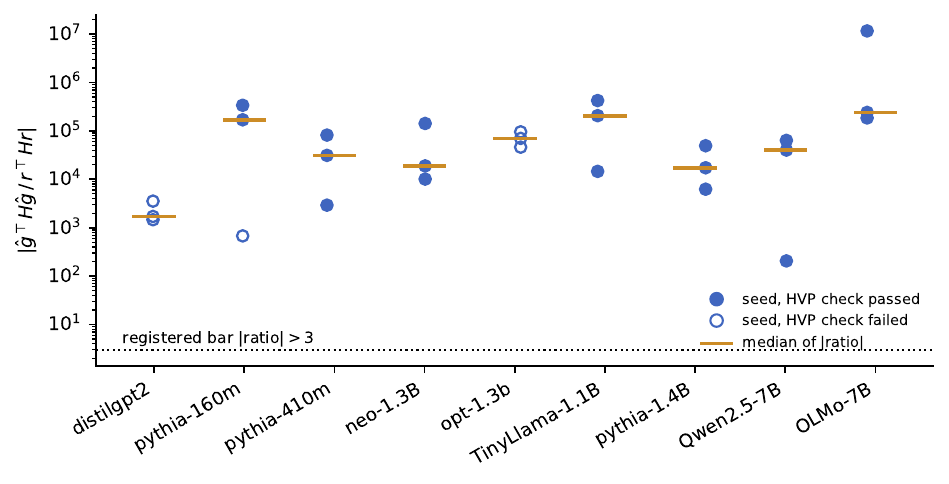}
\caption{\prop{P7} per-seed $|\hat\vg^\top H\hat\vg / \vr^\top H\vr|$; open circles mark seeds failing the finite-difference HVP reconstruction check, amber bars are per-model medians of $|$ratio$|$.}
\label{fig:curvature}
\end{figure}

\paragraph{The mixture-probe curvature ratio is large but denominator-sensitive.}
On the seven model-level cells that pass the finite-difference HVP reconstruction check, \prop{P7}'s absolute ratio to one random Rayleigh quotient ranges from $1.7\cdot10^{4}$ to $2.4\cdot10^{5}$; DistilGPT-2 and OPT-1.3B fail the check (Figure~\ref{fig:curvature}). What this establishes is a large ratio against this particular comparator, and no more than that. A random Rayleigh quotient of an indefinite Hessian can be tiny through sign cancellation, so the ratio does not separate genuine anisotropy from a small sampled denominator; it also varies by orders of magnitude across seeds, and the per-task control below rules out reading it as a task-invariant scalar. A follow-up with $64$ random directions per cell settles the magnitude question: on the seven models passing the reconstruction check, the task-direction ratio to the \emph{median} random quotient is $1.4\cdot10^{4}$--$5.9\cdot10^{5}$, and the more stable comparator $\norm{H\vv}$ gives $71\times$--$590\times$ against its median. The sign is a different matter: the signed task curvature exceeds all $64$ random draws on only four of the seven models ($\vth^\star$ is not a minimum, and the task-direction curvature is negative on some seeds). The task direction is extreme in curvature magnitude, while not always the most positively curved direction.

OLMo-7B-Instruct was chosen as a held-out model because activation steering had transported poorly on it in the companion paper's experiments. It passes every surviving one-direction diagnostic and fails exactly the steering-adjacent \prop{P5}: the one property tied to the activation bridge breaks on the model where that bridge already looked fragile.

\subsection{Controls: the raw base point and per-task probes}
\label{sec:controls}

Two controls pin down what the multitask adaptation actually changes and how robust the diagnostics are to the probe choice.

\paragraph{Is $\vth^\star$ special?}
We repeated half of the battery at the base point (the LoRA adapter set to zero, so the represented function is exactly the pretrained network). The frozen \prop{P7} statistic fails there (only $3/6$ development models pass, none with acceptable across-seed CV), but the single-denominator limitation prevents treating that alone as a signature of priming. More telling is the commutator: the relation $\eta^\dagger\approx0.10/\kappa$ is \emph{already present at base}. The registered scoring finds $38/42$ finite cells, $35/38$ within a factor $1.5$, median multiplicative error $1.099$, Spearman$(\eta^\dagger,\kappa)=-0.994$. Quantitatively, the same cells measured at the base point and at $\vth^\star$ differ by a median factor of $2.6$ in $\eta^\dagger$ and $2.9$ in $\kappa$ (the registered bar for a substantial shift was a factor of $2$). So multitask adaptation shifts the measured scales, while both mechanisms are present already at the base point.

\paragraph{Per-task probes.}
Replacing the mixture probe by each single task (an additional control measured at the base point) leaves \prop{P1} and \prop{P6} stable: no model shows two or more task-level verdict flips, and no model has a task coefficient of variation above $1$. \prop{P7} is the exception (every development model flips on at least two tasks with task CV above $1$), so directional curvature depends on the probe and cannot be read as a task-invariant scalar. This is also why the protocol keeps one fixed mixture estimator for $L$, $\vg$, and $H$ across all properties.

\subsection{The boundaries on one ruler}
\label{sec:ruler}

We registered a prediction that the three ``radii'' (\prop{P1}'s $\sigma^\star_1$, \prop{P4}'s $\alpha^\star_4$, and \prop{P8}'s $\eta^\dagger$) would locate the same scale within a factor of three. As stated, the prediction cannot even be tested: the three live on different axes (unit-direction norm, task-vector fraction, learning rate), so the registered comparison was ill-posed. The conversion matters because learning rates are not comparable across models: a step of $\eta=10^{-3}$ on one model can displace the weights more than a step of $10^{-2}$ on another once the gradient norms differ by an order of magnitude, as they do here. A meaningful comparison therefore needs one ruler, and we use the Euclidean norm of the induced weight-space displacement: a \prop{P1} step of scale $\sigma$ has norm $\sigma$; a \prop{P8} update pair at onset $\eta^\dagger$ displaces the parameters by $r_8=\eta^\dagger\norm{\vg_A+\vg_B}$ to leading order, and both factors are archived per cell; a \prop{P6} draw has norm $\sigma\sqrt P$. (\prop{P4} is the one boundary this conversion cannot yet place: task-vector norms were not archived.)

\begin{figure}[!t]
\centering
\includegraphics[width=0.9\linewidth]{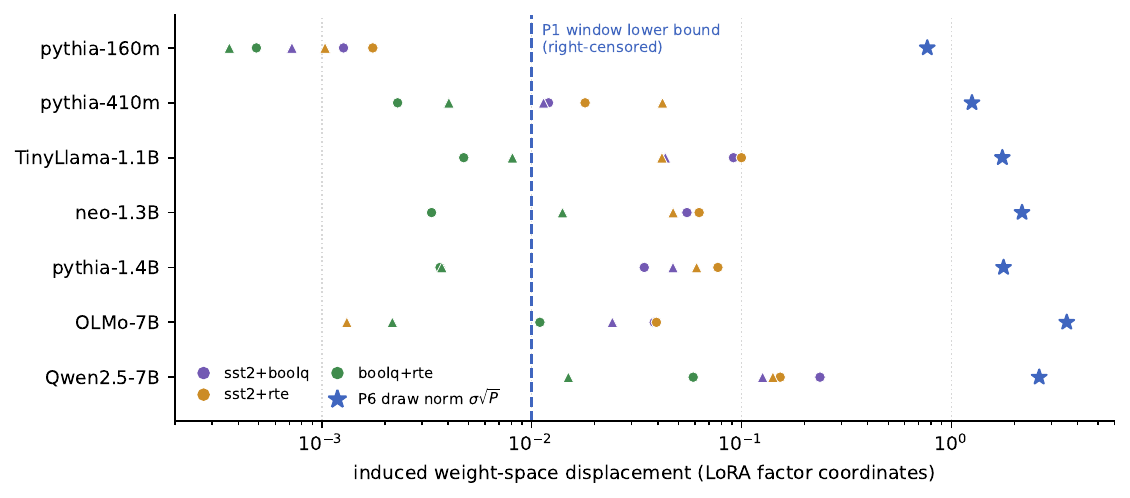}
\caption{All measured boundaries on the common displacement ruler, models ordered by median $\kappa$: per-cell two-direction onsets $r_8=\eta^\dagger\norm{\vg_A+\vg_B}$ (circles seed 0, triangles seed 1), the right-censored one-direction \prop{P1} bound (dashed vertical), and \prop{P6} draw norms $\sigma\sqrt P$ (stars).}
\label{fig:ruler}
\end{figure}

Figure~\ref{fig:ruler} collects every measured boundary on this ruler. On the fine-grid cells of Section~\ref{sec:spine} (seven models, two seeds, three pairs; $41$ measured onsets, one of them left-censored), the two-direction onset displacement spans $r_8\in[3.6\cdot10^{-4},\,0.24]$ across (model, pair) cells, at least a factor of $650$ (the smallest value belongs to the left-censored cell, so its true onset may be smaller still); the pooled median is $1.8\cdot10^{-2}$, and per-model medians run from $8.8\cdot10^{-4}$ (Pythia-160M) to $0.13$ (Qwen2.5-7B). Relative to the one-direction window, whose measured lower bound is $10^{-2}$ on every model, $15$ of $41$ cells break commutativity at displacements strictly \emph{inside} the window (on Pythia-160M the median onset sits $11\times$ inside); for the remaining $26$ cells the relative position is undecided because $\sigma^\star_1$ is right-censored. The \prop{P6} draws measure $0.5$--$3.6$ on the same ruler ($54\times$--$355\times$ the \prop{P1} bound), and first-order scores stay predictive there.

The neighbourhood is thus strongly anisotropic: single-direction moves remain first-order predictable out to displacements one to two orders of magnitude beyond the point where, on over a third of the tested cells, two-direction ordering has already broken. And the two-direction thinness is a per-pair quantity: within one model and seed, the onset can move by an order of magnitude when the task pair changes. The next section asks what sets it.

\section{The Two-Direction Boundary Is a Curvature Commutator}
\label{sec:spine}

The atlas separates single-perturbation predictability from two-update order sensitivity; this section is about the mechanism of the latter. We first derive the leading term, then check how accurately it reproduces the measured defect when both sides are computed from the same minibatches, then push it beyond the LoRA operating point, and finally report a preregistered cross-fitted test of a stronger functional hypothesis, which fails.

As background, the Hessian of a trained network is often described as a low-rank, curved \emph{outlier} subspace plus a high-dimensional, near-flat \emph{bulk}~\citep{sagun2017hessian, papyan2019spectrum}, with different views on where the learning signal lives relative to that split~\citep{gurari2018tiny, song2025tiny, dome2025}. Our own observation below is more narrow: the bracket vector is strongly enriched in the top Hessian subspace relative to the isotropic floor, on the cells that pass the HVP reconstruction check. The bulk/outlier picture serves here as motivation, and we do not study it directly.

\subsection{The leading term}
\label{sec:commutator}

Assume $L_A$ and $L_B$ are $C^3$ in a neighbourhood of $\vth^\star$, and take one gradient step of size $\eta$ on each of two tasks, in the two orders. Expanding the second step around $\vth^\star$,
\begin{align*}
\vDel_{A\to B}(\eta) &= -\eta\,\vg_A - \eta\,\vg_B(\vth^\star-\eta \vg_A) = -\eta(\vg_A+\vg_B) + \eta^2 H_B \vg_A + O(\eta^3),\\
\vDel_{B\to A}(\eta) &= -\eta(\vg_A+\vg_B) + \eta^2 H_A \vg_B + O(\eta^3),
\end{align*}
where $\vg_X=\nabla L_X(\vth^\star)$ and $H_X=\nabla^2 L_X(\vth^\star)$. The first-order parts cancel, leaving the purely second-order commutator
\[
\vDel_{A\to B}-\vDel_{B\to A} = \eta^2\,(H_B \vg_A - H_A \vg_B) + O(\eta^3),
\]
the Lie bracket of the two update fields. In words: whichever task goes second takes its step from ground the first task has already moved, so its gradient has changed by curvature times the first step; if $H_B\vg_A$ and $H_A\vg_B$ differ, the two orders land in different places, and their leading difference is the bracket. When $\norm{\vg_A+\vg_B}>0$, dividing by $\norm{\vDel_{A\to B}}\approx\eta\norm{\vg_A+\vg_B}$ gives the defect to leading order:
\[
c(\eta)\;=\;\eta\,\kappa+O(\eta^2),\qquad \kappa \;=\; \frac{\norm{H_B \vg_A - H_A \vg_B}}{\norm{\vg_A+\vg_B}}.
\]
Two remarks before the measurements. First, the normalization is ill-conditioned when $\vg_A\approx-\vg_B$: a large $\kappa$ can reflect a large bracket numerator, first-order cancellation, or both, and the present tables do not decompose the two. Second, $\kappa=0$ kills the leading term only; higher-order order-dependence can survive. Once the coordinate system, probe estimator, normalization, and defect threshold $0.10$ are fixed, the leading-order onset estimate has no fitted coefficients:
\[
\eta^\dagger\cdot\kappa \;\approx\; 0.10.
\]
The bracket itself and HVP-based order prediction are established in directly overlapping work~\citep{rukhovich2025commute,sweeney2026geometry}, alongside backward-error analyses of sequential updates~\citep{smith2021origin,rosca2021discretization}. What we claim here is only the measurement: the normalized endpoint defect across a model--task matrix, its finite-step calibration in fixed LoRA coordinates, and its comparison with the other atlas diagnostics. The derivation covers memoryless gradient updates; fixed-clock optimizer state can add a leading $O(\eta)$ term~\citep{sweeney2026optimizer}.

\paragraph{The test.}
For each model we measure $\kappa$ from two Hessian-vector products at $\vth^\star$, and we compute the top-$10$ Hessian subspace by Lanczos to ask whether the commutator vector is enriched there relative to the random floor $10/P$. DistilGPT-2 and OPT-1.3B are excluded from all commutator measurements by the \prop{P7} reconstruction check. We made two pre-registered predictions: $\eta^\dagger\cdot\kappa\approx0.10$ within a factor of three, and top-subspace overlap far above the floor. The fine-grid re-measurement protocol below and its metrics were likewise registered before any of its data was read.

\begin{figure}[!t]
\centering
\includegraphics[width=\linewidth]{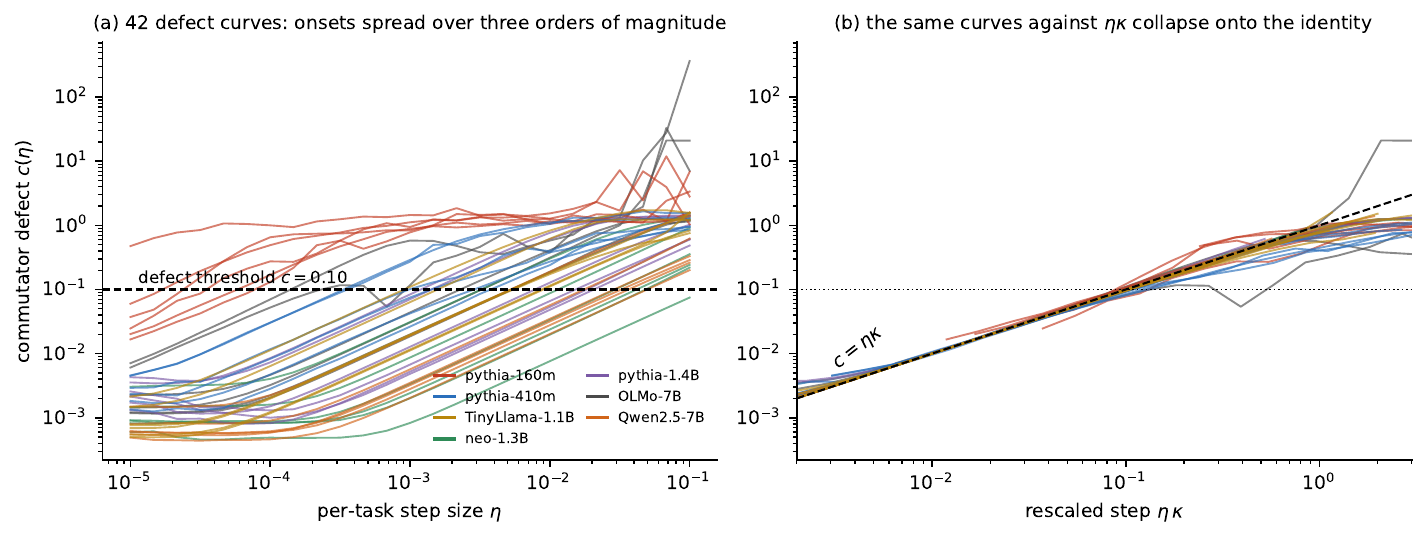}
\caption{Fine-grid defect curves for all $42$ (pair, seed) cells on the seven models passing the HVP reconstruction check: (a)~raw $c(\eta)$; (b)~the same curves against $\eta\kappa$, collapsing onto the identity (dashed).}
\label{fig:defect}
\end{figure}

\begin{table}[!t]
\centering\footnotesize
\setlength{\tabcolsep}{5pt}
\caption{The commutator test, per-model medians over (pair, seed) cells; the product column is the median of per-cell products. $\cos U_{\rm comm}$ is the commutator's top-$10$ Hessian-subspace energy (random floor $10/P$).}
\label{tab:spine}
\begin{tabular}{lrrrr}
\toprule
Model & $\kappa$ & $\eta^\dagger$ & $\eta^\dagger\!\cdot\!\kappa$ & $\cos U_{\rm comm}$ \\
\midrule
pythia-160m & $3710$ & $2.9\cdot10^{-5}$ & $0.096$ & $0.072$ \\
pythia-410m & $98$ & $1.2\cdot10^{-3}$ & $0.109$ & $0.009$ \\
neo-1.3B & $2.9$ & $2.9\cdot10^{-2}$ & $0.099$ & $0.165$ \\
TinyLlama-1.1B & $19.6$ & $5.4\cdot10^{-3}$ & $0.102$ & $0.120$ \\
pythia-1.4B & $9.7$ & $1.2\cdot10^{-2}$ & $0.099$ & $0.061$ \\
Qwen2.5-7B & $3.4$ & $3.1\cdot10^{-2}$ & $0.101$ & $0.013$ \\
OLMo-7B & $37.1$ & $2.8\cdot10^{-3}$ & $0.100$ & $0.028$ \\
\midrule
\textit{median (7 models)} & & & $\mathbf{0.100}$ & \\
\bottomrule
\end{tabular}
\end{table}

\subsection{Verification under a matched estimator}
\label{sec:matched}

Our first measurement, on \prop{P8}'s shared coarse grid with single-batch one-step gradients, gave only an order-of-magnitude law: median $\eta^\dagger\kappa$ at the predicted value but per-model scatter of about $10\times$, with $4$ of $7$ models within the registered factor of three. We then pre-registered and ran a re-measurement designed to remove two possible causes of that scatter: a $25$-point $\eta$ grid from $10^{-5}$ to $10^{-1}$ with the onset read by log-interpolation of the crossing (removing grid quantization and, on all but two cells, censoring), and the same fixed three-batch gradient estimators driving both the two-step defect and the HVP commutator (removing estimator mismatch). Here ``matched'' means that both routes see the same data: the batches that build the two-step paths also build the HVPs, so sampling noise cannot show up as route-to-route disagreement. In this version the scatter collapses. Figure~\ref{fig:defect} shows the measured object itself: every cell's defect curve is a unit-slope line through its own onset, the onsets spread over three orders of magnitude (from $1.7\cdot10^{-5}$ to $5\cdot10^{-2}$), and rescaling each curve by its own $\kappa$ collapses all $42$ onto the identity, which the curves leave only at large defect, where higher-order terms take over. Table~\ref{tab:spine} gives the per-model medians: every product $\eta^\dagger\kappa$ lies in $[0.096,0.109]$, within $9\%$ of the threshold value.

Because $\eta^\dagger$ is read from the same defect curve, $\eta^\dagger\kappa\approx0.10$ is arithmetic once the leading expansion holds pointwise; the substantive content is route-to-route consistency. Away from the threshold, the finite-difference endpoint defect matches the HVP-side leading term $\eta\kappa$ at median ratio $1.002$ (IQR $[0.999,1.013]$), with $94\%$ of $218$ mid-range points within $10\%$. The shared batches make this a stringent numerical check: the expansion is accurate in range, the HVP implementation agrees with finite differences, higher-order terms are small, and float32 effects are controlled. It is not an out-of-sample prediction. The cross-fitted test of Section~\ref{sec:crossfit} measures the out-of-sample version directly, and it fails.

\begin{figure}[!t]
\centering
\includegraphics[width=0.52\linewidth]{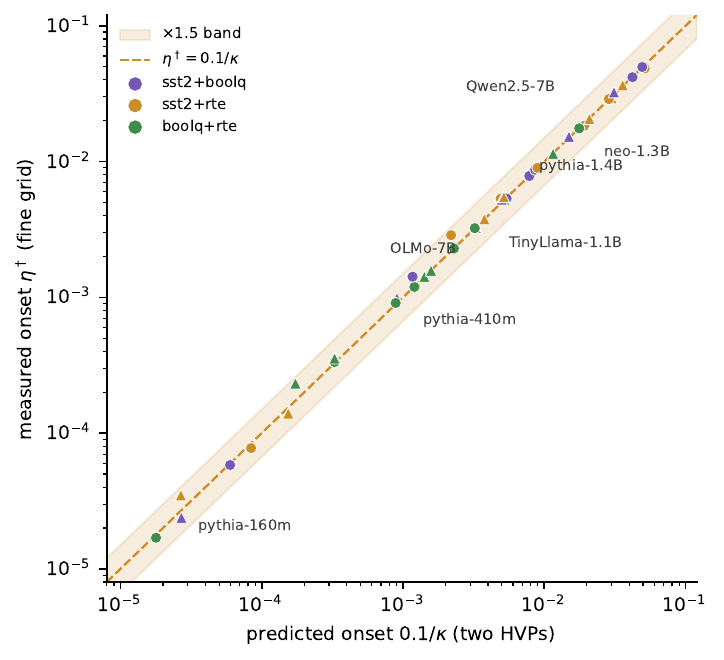}
\caption{Measured onset $\eta^\dagger$ against the two-HVP prediction $0.1/\kappa$ for the $41$ measured fine-grid cells, colored by task pair (circles seed 0, triangles seed 1); the band marks $\times1.5$.}
\label{fig:spine}
\end{figure}

\begin{table}[!t]
\centering\footnotesize
\setlength{\tabcolsep}{5pt}
\caption{Within-seed pair spans of $\kappa$ and $\eta^\dagger$, and per-model prediction errors of $0.10/\kappa$ on the $41$ measured cells.}
\label{tab:perpair}
\begin{tabular}{lrrrr}
\toprule
model & $\kappa$ pair span & $\eta^\dagger$ pair span & median pred./meas. & max pred./meas. \\
\midrule
OLMo-7B & $11.8\times$ & $12.8\times$ & $1.049\times$ & $1.343\times$ \\
Qwen2.5-7B & $4.8\times$ & $4.9\times$ & $1.007\times$ & $1.042\times$ \\
TinyLlama-1.1B & $6.7\times$ & $6.7\times$ & $1.017\times$ & $1.059\times$ \\
neo-1.3B & $9.3\times$ & $8.8\times$ & $1.009\times$ & $1.058\times$ \\
pythia-1.4B & $15.3\times$ & $14.9\times$ & $1.010\times$ & $1.045\times$ \\
pythia-160m & $5.6\times$ & $4.0\times$ & $1.109\times$ & $2.415\times$ \\
pythia-410m & $9.1\times$ & $9.6\times$ & $1.085\times$ & $1.311\times$ \\
\bottomrule
\end{tabular}

\end{table}

The pairwise view carries the most information (Figure~\ref{fig:spine}, Table~\ref{tab:perpair}). Task pairs are not interchangeable: within a single model and seed, $\kappa$ spans up to $15\times$ across the three registered pairs, and the measured onsets move with it; in Figure~\ref{fig:spine} this is visible as the color separation inside each model's cluster. Across the $41$ cells the pooled rank correlation between $\eta^\dagger$ and $\kappa$ is $-0.998$; $36/41$ predictions $\hat\eta=0.10/\kappa$ land within $\times1.2$ of the measured onset and $40/41$ within $\times1.5$ (median multiplicative error $1.020$), with the single outlier left-censored at the grid edge, so its apparent factor-$2.4$ error is conservative. Pair-to-pair variation therefore \emph{strengthens} the result: some task pairs commute longer than others because their cross-curvature is smaller, and two HVPs suffice to know which is which.

A registered decomposition run settles where the original scatter came from. Re-measuring on the fine grid but with the \emph{original} single-batch defect estimators (HVP side unchanged), the per-cell scatter returns to the coarse level ($0.26$, versus $0.009$ matched and $0.24$ coarse): grid quantization contributed almost nothing, and the scatter was essentially all minibatch noise of the probe. Even then the median-level prediction survives (all seven models' median products land within a factor of $1.7$ of $0.10$, at $0.061$--$0.161$), so the two-HVP commutator gives a useful warning signal even with a noisy single-batch probe, while per-instance predictions inherit the probe's variance. The second registered prediction also holds: on all seven models the commutator vector's top-$10$ Hessian eigenspace overlap is three to five orders of magnitude above the $10/P$ floor (the lowest cell is $637\times$ floor), although the absolute top-$10$ energy can still be small (Table~\ref{tab:spine}). We also tested whether the \emph{degree} of top-eigenspace engagement modulates the onset beyond $\kappa$ itself, and found nothing (partial rank correlation $+0.000$ controlling for $\kappa$), though with the pooled correlation at $-0.998$ there is almost no residual variance left for a second predictor.

\subsection{Beyond the LoRA operating point: full fine-tuning and 13--14B scale}
\label{sec:beyond}

Everything above lives at a LoRA operating point. A registered control repeats the fine-grid protocol with $\vth^\star$ trained by \emph{full} fine-tuning (all parameters, lr $2\cdot10^{-5}$, same recipe) on the three models small enough for full-parameter Hessian work (Pythia-160M/410M, TinyLlama-1.1B), with the grid extended to $10^{-7}$ because full-space commutators are much larger. That extension is a finding in itself: $\kappa_{\rm full}/\kappa_{\rm LoRA}$ has median $4$ and reaches $60$, so full-fine-tuning order sensitivity sets in earlier than matched LoRA order sensitivity. The threshold-level product carries over at the median: $\eta^\dagger\kappa=0.090$, $0.097$, $0.116$ on the three models, with the $18$ finite cell products spanning $0.068$--$0.217$, no censoring, and $94\%$ single crossings; the commutator stays aligned with sharp Hessian directions (model-median overlap at least $10^{5}\times$ the floor, with $P$ now the full parameter count). The pointwise check away from the threshold does \emph{not} pass this control: the measured-to-predicted ratio there is $1.56$, outside the registered $[0.9,1.1]$ band. We believe this failure is instrumental: the band was calibrated at LoRA-scale $\kappa$, and at full-FT $\kappa$ it emphasizes very small $\eta$ near the float32 arithmetic floor (the pooled measured/predicted ratio is largest at the smallest grid values and settles near one by $\eta=10^{-4}$). Still, the registered check failed. At the crossing itself the signal dominates the floor, which is why the threshold products are unaffected.

We also extended the LoRA protocol to larger base models, Qwen2.5-14B and OLMo-2-13B~\citep{olmo2team2024olmo2}. This gives six finite observed crossings with $\eta^\dagger\kappa\in[0.098,0.103]$ across the two models and two seeds; the remaining six pairs are right-censored at the upper grid edge, so this is scale-consistency evidence on the finite cells, and the censored pairs remain undecided. (A third registered scale cell, Llama-3.1-Tulu-3-8B, was dropped before scoring after a pinned-environment tokenizer failure; no Tulu number is reported.)

\subsection{A cross-fitted functional forecast fails}
\label{sec:crossfit}

The matched experiment asks whether two routes agree when they share estimators. We separately preregistered the stronger claim a practitioner would actually want: a bracket estimated on one sample should predict an endpoint built from a second sample and signed task effects evaluated on a third. This three-sample design mimics how the diagnostic would actually be used: geometry estimated today, updates built from tomorrow's batches, effects checked on data neither step has seen. For geometry sample $G$, update sample $U$, and evaluation sample $E$, the registered quantities are
\[
\begin{aligned}
\vb_G&=H_B^G\vg_A^G-H_A^G\vg_B^G,
&\vq_U&=\frac{\vth_{A\to B}^{U}-\vth_{B\to A}^{U}}{\eta^2},\\
\widehat y_C&=(\vg_C^G)^\top \vb_G,
&y_C&=\frac{L_C^E(\vth_{A\to B}^{U})-L_C^E(\vth_{B\to A}^{U})}{\eta^2},
\end{aligned}
\]
tested on four new confirmation models (GPT-2-large, OLMo-2-7B, Mistral-7B-v0.3, Qwen2.5-14B), three seeds, all ten unordered pairs of the five tasks, five evaluation tasks per pair, with disjoint $60$/$60$/$128$-example blocks per task and the step $\eta=0.05/\kappa_G$ clipped to $[10^{-5},3\cdot10^{-2}]$ (clipped cells are excluded from the primary aggregates). The registered rules required absolute predictive power and step-size eligibility; a favorable pair-specificity comparison could not rescue a failure against the zero baseline. All twelve runs passed a frozen check of the raw outputs before the single scoring pass.

\begin{table}[!t]
\centering\small
\setlength{\tabcolsep}{5pt}
\caption{The preregistered cross-fitted confirmation at a glance.}
\label{tab:crossfit}
\begin{tabular}{p{0.29\linewidth}p{0.52\linewidth}p{0.12\linewidth}}
\toprule
Registered component & Fresh confirmation result & Verdict \\
\midrule
Raw-output validity & $12/12$ runs; $1800$ scalar checks pass & pass \\
Instrument eligibility & $73/120$ pair cells limited ($60.8\%>25\%$) & invalid \\
Prediction $R^2$ against zero & $R^2_{\rm zero}=-560.419$; $1/4$ models positive & fail \\
Endpoint direction & median cosine $0.0429$; required $>0.50$ & fail \\
Pair specificity & beats shuffled pairs: $R^2_{\rm zero}>-561.470$, cosine $>0.00433$ & pass \\
Same-sample diagnostic & endpoint cosine $0.9987$; functional $R^2_{\rm zero}=-21.337$ & diagnostic \\
Primary forecast & three required bars fail & not validated \\
\bottomrule
\end{tabular}
\end{table}

The forecast fails, in two ways (Table~\ref{tab:crossfit}). The first is a matter of step size: only $47/120$ primary pair cells stay inside the declared interval (these are the \emph{eligible} cells), so the preregistered rule calls the aggregate invalid (it is not a clean confirmatory negative either): the registered step scale is infeasible for most (model, pair) cells. But conditional on the eligible cells, both absolute prediction bars also miss by wide margins: $R^2_{\rm zero}=-560.4$ (the explained variance of the fixed prediction, measured against the zero baseline; negative values mean that predicting zero would have been better) and median endpoint cosine $0.043$. Scoring each prediction against its own pair does slightly better than scoring it against a cyclically shuffled pair (a control in which every prediction is matched with the wrong pair's measured effect), so the bracket carries \emph{some} pair-specific signal, but neither beats the zero-effect baseline. The same-sample diagnostic sharpens the picture: when geometry and update come from the same sample, the bracket reconstructs the endpoint direction almost exactly (cosine $0.9987$), while across samples the median endpoint cosine is $0.043$; even the endpoint geometry fails to transfer between update samples, before any functional question arises, and the matched functional score is negative as well ($R^2_{\rm zero}=-21.3$). The narrow conclusion is therefore twofold: the matched Lie-bracket calculation is a strong endpoint Taylor check, and this cross-fitted implementation supplies no evidence for a sample-stable predictor of either endpoints or effects. Why the registered step scale is so often infeasible, and whether any signal survives better step policies, is the next question this line of work has to answer.

\subsection{An exploratory association, and a forecast that failed}
\label{sec:lammax}

\begin{figure}[!t]
\centering
\includegraphics[width=0.56\linewidth]{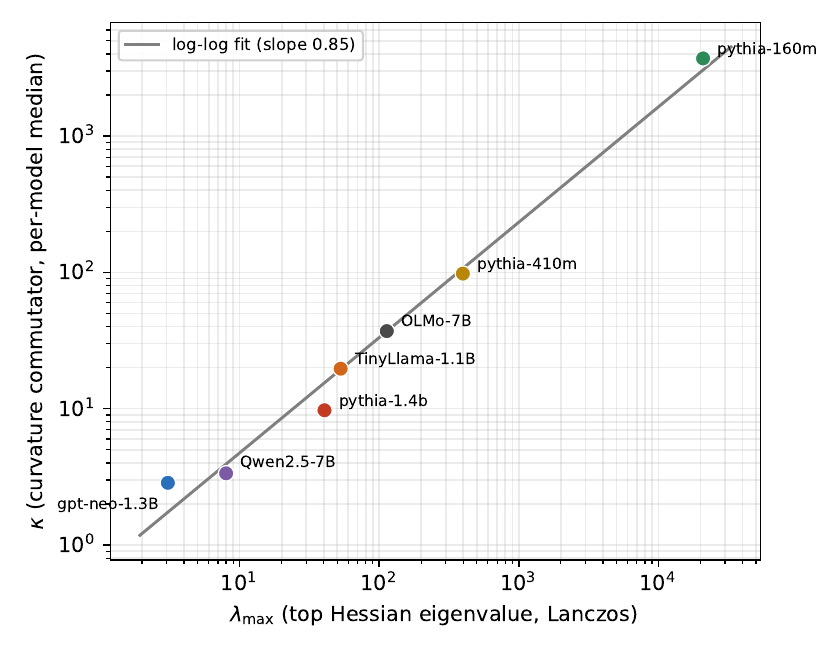}
\caption{Model-median $\kappa$ against the probe-loss $\lambda_{\max}$ on the seven models passing the HVP reconstruction check (log--log).}
\label{fig:lamkappa}
\end{figure}

Across the seven model medians that pass the HVP reconstruction check, $\kappa$ is rank-associated with the probe-loss $\lambda_{\max}$ at Spearman $+1.0$ with log--log slope $0.85$ (Figure~\ref{fig:lamkappa}). This looks like a useful screening heuristic: one Lanczos $\lambda_{\max}$ estimate per model instead of two HVPs per pair. It is, however, a seven-point, within-atlas association with model-median $\kappa$; it shares probe data with the quantity it predicts; it omits the pair-specific variation of Table~\ref{tab:perpair}; the scoring script's $\lambda_{\max}$ reproduction check flagged one seed-level mismatch (Pythia-160M seed 1: $5972.66$ vs $6372.72$, just outside the $5\%$ tolerance); and the Hutchinson trace estimator failed its registered stability check at both $8$ and $64$ probes (CV above $0.5$ on $10/14$ cells, negative mean traces persisting on two models), so we cannot distinguish outlier summaries from bulk-inclusive curvature. At an operating point that is not a loss minimum, the trace can be a small signed difference of large positive and negative spectral mass; a bulk-only or absolute-value summary is needed, and we leave it open. The $13$--$14$B cells are an exploratory overlay only: their finite threshold products stay in $[0.098,0.103]$, but their $\kappa$ values sit up to $\times5.5$ below the seven-model fit, so they do not strengthen the association.

We then froze the seven-model fit and applied it once, prospectively, to three new models (GPT-2-medium, OPT-350M, Qwen2.5-1.5B). The forecast did not validate. Qwen2.5-1.5B had a finite onset within $\times1.28$ of the forecast, but the other two models were left-censored at the lower grid edge, leaving the registered radius verdict undecidable; and the paired comparison of first-order score correlations inside versus outside the forecast radius failed outright (GPT-2-medium $0.882$ vs $0.871$; Qwen2.5-1.5B $0.999$ vs $1.000$). The one-direction/two-direction explanation for that null is post hoc. We therefore retain $\lambda_{\max}$ as an exploratory within-atlas association; the deployable local diagnostic remains the pair-specific two-HVP $\kappa$, within the matched-protocol scope established above.

\section{Descriptive Headroom and Search-Scale Checks}
\label{sec:headroom}

The atlas established that single perturbations are first-order predictable; this section quantifies what that gives, in nats and in search terms. Both measurements here are descriptive reanalyses of archived measurement records (no thresholds, no selection), and neither establishes value created by multitask priming. Throughout, \emph{headroom} is the measured reduction in the probe's token-mean cross-entropy (nats per active continuation token) along one direction over the tested grid $\sigma\le10^{-2}$. On all but one (model, task) cell the \prop{P1} window is right-censored at that edge, so those values are also within-window lower bounds; the exception is ARC-Easy on Pythia-160M, whose median window ends at $3\cdot10^{-3}$, inside the grid.

\begin{figure}[!t]
\centering
\includegraphics[width=0.82\linewidth]{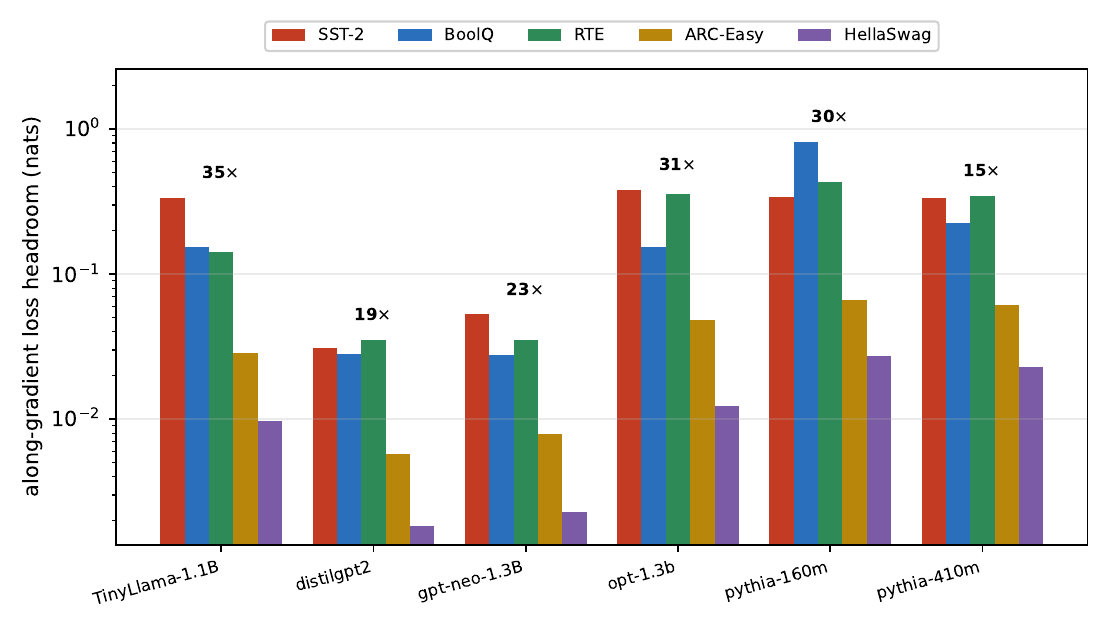}
\caption{Base-point per-task along-gradient loss reduction over the tested \prop{P1} grid (median over three seeds, log axis); right-censored on all but one cell (see text).}
\label{fig:headroom}
\end{figure}

\paragraph{Per-task headroom is a gradient-norm map, with a wide task span.}
Table~\ref{tab:headroom} and Figure~\ref{fig:headroom} report, for each development model and task, the loss reduction along that task's own gradient over the tested grid, at the raw base checkpoint where the per-task battery ran (median over three seeds). Across cells the values range from $0.002$ to $1.0$ nats per token ($0.05\%$--$20\%$ of the corresponding probe loss), and the within-model task span is $15$--$35\times$. Within a shared window this ordering is just the per-task gradient-norm ordering, which is what makes it predictable in advance: to first order the reachable reduction is slope times window width, the slope is the gradient norm, and the window is shared, so ranking the tasks costs one gradient evaluation each. Two scoping notes temper it: continuation lengths, tokenizations, and choice counts differ across tasks, so raw cross-entropy is not a common functional currency; and the table is measured at the raw base checkpoint. At $\vth^\star$ itself, the mixture-probe headroom is $0.007$--$0.047$ nats per token ($0.2\%$--$2.0\%$ of probe loss) across the nine models: residual signal exists at the operating point, but per-task value there was not measured.

\begin{table}[!t]
\centering\footnotesize
\setlength{\tabcolsep}{5pt}
\caption{Per-task along-gradient headroom (nats per active continuation token) at the raw base checkpoint, median over three seeds; ``span'' is the max-to-min task ratio. Values are measured over the full tested grid, right-censored on all but one cell (see text).}
\label{tab:headroom}
\begin{tabular}{lrrrrrr}
\toprule
model & \task{sst2} & \task{boolq} & \task{rte} & \task{arc-e} & \task{hella} & span \\
\midrule
TinyLlama-1.1B & $0.335$ & $0.154$ & $0.143$ & $0.029$ & $0.010$ & $35\times$ \\
distilgpt2 & $0.031$ & $0.028$ & $0.035$ & $0.006$ & $0.002$ & $19\times$ \\
neo-1.3B & $0.053$ & $0.028$ & $0.035$ & $0.008$ & $0.002$ & $23\times$ \\
opt-1.3b & $0.383$ & $0.154$ & $0.356$ & $0.048$ & $0.012$ & $31\times$ \\
pythia-160m & $0.341$ & $0.818$ & $0.433$ & $0.067$ & $0.027$ & $30\times$ \\
pythia-410m & $0.333$ & $0.225$ & $0.346$ & $0.061$ & $0.023$ & $15\times$ \\
\bottomrule
\end{tabular}

\end{table}

\paragraph{Two best-of-$256$ runs land near the Gaussian order-statistic scale.}
Under \prop{P6} the same $32$ examples define both $\vg$ and the realized gain, so the high correlations are in-sample; what we can check descriptively is the \emph{magnitude} that best-of-$N$ search extracts. For an isotropic draw of fixed radius $r$, the first-order gain is $-\vg^\top\vdel$ with normalized projection approximately $\mathcal N(0,1/P)$, so the leading expected best-of-$N$ gain is
\[
\Delta L_{\text{best-of-}N}\;\approx\;\norm{\vg}\,r\,\sqrt{\tfrac{2\ln N}{P}},
\]
the fixed-radius specialization of the companion theorem~\citep[Theorem~1]{piontkovskaia2026recoverable}. At $N=256$ the leading term overstates the exact standard-normal expected maximum ($2.8269$ vs $\sqrt{2\ln 256}=3.3302$, about $18\%$ high), so we correct by that factor. On GPT-2-medium and Qwen2.5-1.5B the search samples on the sphere of radius $r=\eta_{\rm forecast}\norm{\vg}$; the realized best-of-$256$ \task{boolq} gains are $2.8\cdot10^{-4}$ and $2.2\cdot10^{-4}$ nats per token against predictions of $3.5\cdot10^{-4}$ and $2.6\cdot10^{-4}$, so the realized-to-predicted ratios are about $0.79$ and $0.85$. These two runs are descriptive; a calibration law would need multiple $N$, radii, and seeds. A rerun of both cells with accuracy recorded answers the conversion question at these radii: the selected perturbations leave multiple-choice accuracy exactly unchanged, on the target and on all four side tasks, on both models, at every multiplier up to $4\times$ the forecast radius ($112$ and $118$ usable target examples). At gains of $\sim2\cdot10^{-4}$ nats, the search headroom is real in loss and invisible in accuracy. The more important content is the $1/\sqrt P$ factor itself: at fixed radius, undirected search receives only a $\sqrt{2\ln N/P}$ share of the along-gradient ceiling, about $0.2\%$ here. First-order scoring is therefore most relevant for constrained or externally generated proposal libraries: block-restricted edits, side-task-filtered candidates, discrete proposal sets. When an unconstrained gradient step is available, it wins. Cross-fitted scores and held-out selection regret remain unmeasured.

\section{What the Atlas Means in Practice}
\label{sec:practical}

No single safe step size falls out of the atlas. What it gives a practitioner is a way to decide which local approximation a given adaptation tool is actually asking for, and what evidence exists for it.

\paragraph{Match the diagnostic to the operation.}
For one intended direction, \prop{P1} checks probe-loss antisymmetry over a scale grid, and \prop{P6} checks first-order ranking at its tested proposal scale on the same probe. Together they are the reason local random search is usable here: linearity makes a proposal's gain equal to its gradient projection, so candidates can be scored and ranked before any of them is run, and best-of-$N$ then amplifies the best signal. For two \emph{sequential memoryless gradient steps}, the pair-specific bracket $\kappa$ gives the leading coordinate-space order defect, with $\eta^\dagger\approx0.10/\kappa$ under the shared-minibatch protocol of Section~\ref{sec:matched}; two HVPs per pair are enough for the warning signal, and pair-to-pair differences are large enough to matter. Simultaneous adapter merging is a different second-order object: \prop{P4} supplies a narrow activation-probe diagnostic (with error growing $\propto\alpha$), and this paper has no validated commutator-based merge-radius calculator.

\paragraph{What not to use as a marker.}
The registered rank-$\le32$ mixture-covariance test was invalid, and its control shows no plateau through $m=256$: do not read mixture-gradient rank at small sample counts as evidence of low-dimensional adaptation. The global mean-vector \prop{P5} bridge has no passing model median: do not treat weight updates and CAA mean vectors as interchangeable, even though input-conditioned correspondences remain open. The measured BoolQ tangent subspace rotates within tens of steps: do not carry a task subspace across a trajectory without rechecking it. Each of these failures is specific to one cheap construction; stronger versions (per-task covariance estimates, input-conditioned correspondences, refreshed subspaces) remain untested.

\paragraph{The bracket's record.}
Endpoint consistency with shared minibatches: verified (median ratio $1.002$). Model-median threshold products across LoRA, full fine-tuning, and $13$--$14$B scale: $0.090$--$0.116$, with full-FT cell-level products spanning $0.07$--$0.22$. Out-of-sample $\lambda_{\max}$ radius forecast: undecidable on two of three models, failed inside/outside search comparison. Cross-fitted prediction on new models: aggregate invalid under the step-size rule, eligible subset strongly negative. The pattern is consistent: the local, matched-estimator statements hold, and every transferable one has so far failed. A practitioner should use $\kappa$ as a cheap, pair-specific, same-probe order-sensitivity diagnostic, and should not use it as a cross-sample effect-size predictor.

\section{Discussion and Limitations}
\label{sec:discussion}

\paragraph{Empirical conclusions.}
Individual perturbations remain first-order predictable on the measured probe, through the tested grid on every model median, while interactions between task updates produce substantial, pair-specific order effects at scales that can sit well inside the one-direction window; this scale gap is the robust organizing fact of the atlas. The interactions decompose cleanly by symmetry: self-curvature governs one-direction antisymmetry (\prop{P1}, \prop{P7}), a symmetric mixed derivative governs simultaneous activation composition (\prop{P4}, with its predicted $\propto\alpha$ error growth visible in the data), and the antisymmetric Lie bracket governs memoryless update order (\prop{P8}), with no shared scale among the three. For the bracket, our contribution is the normalized finite-step atlas in fixed LoRA coordinates: the law $c(\eta)=\eta\kappa$ holds pointwise at median ratio $1.002$ when both sides share minibatches, the onsets it predicts span three orders of magnitude across (model, pair) cells, and the median threshold products persist at full fine-tuning and $13$--$14$B scale. The stronger hypotheses we tested did not survive: the disjoint-sample forecast is invalid under its step-size rule with a strongly negative eligible subset, and the $\lambda_{\max}$ radius forecast was undecidable on two of its three models, with the inside/outside search comparison failing as well. We call $\vth^\star$ \emph{primed} only to record that it has seen the mixture and retains residual gradients; the base-point controls show the mechanisms are present without it.

\paragraph{Limitations.}
The most important limitations of our study are as follows.
\begin{enumerate}
\item \emph{Coordinate and optimizer scope.} LoRA factor coordinates have gauge freedom, and every Euclidean parameter-space quantity in this paper is gauge-dependent. The bracket expansion describes memoryless gradient steps; fixed-clock momentum or AdamW state can introduce $O(\eta)$ order effects. No gauge-rescaling, balanced-gauge, Fisher, or output-KL control is yet reported (the effective-weight control for \prop{P4} now is; Section~\ref{sec:failures}).
\item \emph{Operating point and probe.} The battery is measured at one LoRA operating point on a five-task classification mixture, not open-ended generation; \prop{P3} is a BoolQ single-task trajectory, and \prop{P4}/\prop{P8} use three registered pairs among \task{sst2}, \task{boolq}, \task{rte}, so the rotation and two-direction claims should be read at that scope. The extended sweeps and their random-direction comparison cover three models and one seed.
\item \emph{Estimator and cross-fit scope.} \prop{P6} scores and evaluates on the same $32$ examples, and the $1.002$ commutator ratio uses matched batches on both routes. The cross-fitted confirmation avoids this overlap but is formally invalid because $60.8\%$ of primary cells require step clipping; its eligible subset is strongly negative. The matched result is an implementation/Taylor check; the cross-fit supplies no positive claim about transfer.
\item \emph{Comparator scope.} \prop{P7}'s original battery used one random Rayleigh quotient per seed; the $64$-direction follow-up confirms the magnitude ratios and the $H^2$ comparator, so what remains open is the probe itself: the ratio is a property of the mixture probe, and the per-task control shows it is not task-invariant.
\item \emph{Scale of the commutator extensions.} The full-fine-tuning control is a $\le1.1$B result whose median threshold products carry over while its pointwise check fails (plausibly for float32 reasons), and the $13$--$14$B cells are six finite crossings plus six right-censored pairs, with the registered Tulu-8B cell dropped for an environment failure; there is no full atlas beyond $7$B.
\item \emph{Registration is partial.} The property list, split, keep rule, and thresholds were registered, but the operating point, estimator details, and \prop{P7}'s final phrasing were frozen in code rather than in the dated document, and several passing statistics fail the registered CV clause (Appendix~\ref{app:frozen}).
\item \emph{Forecast negatives.} The out-of-sample $\lambda_{\max}$ radius forecast was undecidable (two of three models left-censored), the inside/outside search comparison failed, and the cross-fitted confirmation was invalid under its step-size rule. The commutator analysis is neither a radius calculator nor a held-out task-effect predictor.
\item \emph{Headroom measurement gaps.} The per-task headroom table is base-point data, accuracy was recorded only in the three-model extended sweeps and the best-of-$256$ rerun, and the \prop{P4} boundary cannot be placed on the common displacement ruler because task-vector norms were not archived.
\item \emph{The functional probes are narrow.} \prop{P4} uses one layer, one mean last-token activation summary, $16$ fixed SST-2 prompts, one task-vector recipe, and three pairs; \prop{P5} one SST-2 global mean-vector construction at one step size over three candidate layers; the showcase accuracy uses $55$--$64$ examples per task (full per-cell table in Appendix~\ref{app:showcaseacc}). Headline words such as ``additivity'' and ``correspondence'' should be read at this width, and \prop{P6} remains in-sample even after the scale sweep: the same probe defines the scores and the gains.
\end{enumerate}

\section{Conclusion}
\label{sec:conclusion}

In this paper, we set out to turn a familiar intuition, that fine-tuning operates in a locally linear neighbourhood, into measured falsifiable statements. As a result, we have found interesting structure but no single radius. On the positive side, we have found a searchable regime: around one multitask LoRA operating point, measured identically on nine models, probe loss changes stay first-order dominated through the atlas grid on every model median, and local perturbations keep the ranks of their gradient projection. In our extended sweeps (three models, one seed), we have found both the interior and the end of the window, and the response along two fixed random directions is $480\times$--$4300\times$ weaker at the edge. 

On the other hand, the same neighbourhood is narrow and pair-specific for interactions: task-gradient subspaces rotate within tens of steps, additivity under the selected activation probe fails at full task-vector scale on several models including both held-out $7$B models, and reordering two task updates produces defects whose onset varies by three orders of magnitude across (model, pair) cells. Several registered claims did not survive their own controls: the rank-$\le32$ mixture-covariance bar was algebraically non-falsifiable and its follow-up showed no plateau, and the global mean-vector weight-to-CAA correspondence has no passing model median.

The two-direction boundary has a classical mechanism: the Lie bracket of the two tasks' gradient vector fields. In fixed LoRA coordinates, the normalized defect follows $c(\eta)=\eta\kappa+O(\eta^2)$ so precisely when both sides share minibatches (median ratio $1.002$) that the threshold law $\eta^\dagger\approx0.10/\kappa$ turns into an actual arithmetic dependency. The two Hessian-vector products measure, pair by pair, a quantity that finite-step experiments then confirm across LoRA, full fine-tuning, and $13$--$14$B scale at the median-product level. The bracket, however, does not yet transfer: estimated on one sample, it fails to predict either the endpoint direction or the task effects of updates built and evaluated on other samples, partly because the registered step scale is infeasible for most cells and partly because the eligible cells show no predictive power. We consider finding out why (step policies, gauge and function-space metrics, optimizer state, or something more fundamental about minibatch geometry) the most important open question that this work leaves.

The broader picture looks the same way as in the companion paper~\citep{piontkovskaia2026recoverable}. There, we found that linear structure in trained networks is real but local, and now we can add: its validity is operation-specific. Self-curvature, symmetric composition, and antisymmetric order effects are different tensors with different scales, and treating ``the linear regime'' as one number conflates them. The practical corollary is modest but usable: check the diagnostic that matches your operation, at your scale, on your pair. The natural theoretical next step is to make the local calculus of adaptation coordinate-free: in further work we hope to study gauge-invariant adapter geometry, optimizer-aware brackets in augmented state space, and function space metrics. To sum up, a shared first-order window makes local random search predictable, while pairwise task interactions often become fragile even at much smaller, pair-specific scales.

\bibliographystyle{plainnat}
\bibliography{references}

\clearpage

\appendix

\section{Registration and protocol provenance}
\label{app:frozen}
The property list, development/held-out split, keep bar ($\ge4/6$ development models with across-seed coefficient of variation below $0.25$, then $\ge2/3$ held-out), and the property thresholds were registered before the sweep, with the following qualifications.

\begin{table}[!t]
\centering\small
\caption{Analysis provenance; ``registered'' applies to the listed object, not to every implementation detail or later interpretation.}
\label{tab:provenance}
\begin{tabular}{p{0.49\linewidth}p{0.43\linewidth}}
\toprule
Analysis object & Status \\
\midrule
Property list, numeric bars, model split, keep/drop rule & Prospectively registered \\
Operating point and estimator details & Frozen in executable code; not fully specified in the dated registration \\
Fine-grid commutator and estimator-decomposition studies & Separately registered before their data were read \\
Base/per-task, full-FT, scale, $m$-sweep, and forecast controls & Separately registered before their data were read \\
Cross-fitted functional commutator confirmation & Separately preregistered; complete raw audit, then a single scoring pass \\
Four follow-up controls (\prop{P7} distribution, \prop{P6} scale sweep, \prop{P4} effective-weight addition, best-of-$N$ accuracy) & Separately preregistered with numeric outcome bars before launch \\
Headroom and common-displacement conversion & Descriptive reanalysis of archived measurement records \\
$\lambda_{\max}$ relation & Exploratory within-atlas association followed by one frozen out-of-sample test \\
\bottomrule
\end{tabular}
\end{table}

\emph{Apparatus.} The original preregistration document retained two stale apparatus phrases from an earlier harness revision (a ``60-step snapshotted recovery run'' and \prop{P2} as a ``snapshotted update covariance''). The executable harness used for every sweep fixes the actual recipe of Section~\ref{sec:properties} (a $300$-step multitask LoRA operating point and per-minibatch gradient covariance), and every run's output records it. Because there is no dated pre-data amendment for the apparatus wording, we do not claim a fully preregistered apparatus: the binding registered objects are the property list, split, keep/drop rule, and thresholds; the operating point and estimator details are frozen-code protocol choices.

\emph{Gating.} After the development sweep we froze the keep/drop decisions in a dated file that applied the threshold bars and recorded the CV qualification separately; several statistics that passed their bars do not satisfy the original CV clause, and we report them as qualified. The frozen decisions were: keep \prop{P1}, \prop{P2}, \prop{P3}, \prop{P4}, \prop{P6}, \prop{P7}, \prop{P8}; drop \prop{P5}. Two registered decisions are worth restating: \prop{P3} was expected to \emph{fail} its naive stable form and was kept as the rotation finding; and the prediction that the three radii coincide was registered and is reported as falsified. For \prop{P7}, the original document specified directional curvature with a reconstruction gate but not the final magnitude-ratio phrasing; the frozen held-out decision file fixed $|\text{ratio}|>3$ as the operative statistic, with cells failing the reconstruction check marked but not counted.

\emph{Later additions.} The fine-grid commutator re-measurement, its matched-estimator metrics, the estimator-decomposition run, the full-fine-tuning control, the $m$-sweep control, the base-point control, the per-task-probe control, the $\lambda_{\max}$ predictor analysis, the out-of-sample forecast/search test of Section~\ref{sec:lammax}, the cross-fitted functional confirmation of Section~\ref{sec:crossfit}, and the showcase run of Section~\ref{sec:firstlook} were each pre-registered before their data was read, in dated files accompanying the respective runs. The headroom analysis (Section~\ref{sec:headroom}), the common-ruler conversion (Section~\ref{sec:ruler}), and the per-cell showcase accuracy table (Appendix~\ref{app:showcaseacc}) are descriptive reanalyses of already-archived outputs by pinned scripts, with no new thresholds and no selection. The four follow-up controls (the \prop{P7} direction distribution, the \prop{P6} scale sweep, the \prop{P4} effective-weight addition, and the best-of-$N$ accuracy rerun) were preregistered with numeric outcome bars before launch; one of their ten bars missed (the signed-extremeness bar for \prop{P7}), and it is reported as such in Section~\ref{sec:survivors}. A later preregistered task-conditioned covariance follow-up stopped at a failed whole-matrix numerical audit before scientific scoring; no partial metrics were used.

\emph{Seed-level facts the medians hide.} One Pythia-160M seed has $\sigma^\star_1=10^{-4}$; eight of $27$ produced \prop{P1} development seed curves are nonmonotone at the tiniest scales, so ``censored on every model'' is a largest-passing-grid median statement; and the \prop{P7} ratio varies by over two orders of magnitude across seeds within some models (the absolute ratios remain well over the magnitude threshold, but clean evidence still requires the instrument gate). Appendix~\ref{app:seeds} lists every per-seed value.

\section{Experimental details}
\label{app:details}
All details below are read from the frozen measurement harness; the archived per-run output records the same configuration.

\paragraph{Models.}
\sloppy Development: \path{distilgpt2}, \path{EleutherAI/pythia-160m}, \path{EleutherAI/pythia-410m}, \path{EleutherAI/gpt-neo-1.3B}, \path{facebook/opt-1.3b}, \path{TinyLlama/TinyLlama-1.1B-intermediate-step-1431k-3T}. Held-out: \path{EleutherAI/pythia-1.4b}, \path{Qwen/Qwen2.5-7B}, \path{allenai/OLMo-7B-Instruct-hf}. Commutator scale extension: \path{Qwen/Qwen2.5-14B}, \path{allenai/OLMo-2-1124-13B} (the registered \path{allenai/Llama-3.1-Tulu-3-8B} cell was dropped before scoring for a pinned-environment tokenizer failure). Full-fine-tuning control: Pythia-160M/410M and TinyLlama-1.1B. Out-of-sample forecast/search models (Sections~\ref{sec:lammax} and~\ref{sec:headroom}): \path{gpt2-medium}, \path{facebook/opt-350m}, \path{Qwen/Qwen2.5-1.5B}. Extended sweeps (Section~\ref{sec:firstlook}): \path{EleutherAI/pythia-410m}, \path{EleutherAI/gpt-neo-1.3B}, \path{Qwen/Qwen2.5-7B}, one seed. Cross-fitted confirmation (Section~\ref{sec:crossfit}): \path{gpt2-large}, \path{allenai/OLMo-2-1124-7B}, \path{mistralai/Mistral-7B-v0.3}, \path{Qwen/Qwen2.5-14B}.

\paragraph{Operating point and LoRA configuration.}
LoRA rank $16$, $\alpha=32$, dropout $0$, no bias terms, applied to the attention Q/K/V input projections in each family's naming: \texttt{c\_attn} (fused QKV) for the GPT-2 family, \texttt{query\_key\_value} (fused) for Pythia/NeoX, and \texttt{q\_proj}/\texttt{k\_proj}/\texttt{v\_proj} for GPT-Neo, OPT, Qwen, TinyLlama, and OLMo; output projections are not adapted. Trainable parameter counts $P$ range from $\sim1.6$M (GPT-2-medium) through $4{,}718{,}592$ (GPT-Neo-1.3B) to $\sim10^{7}$ at $7$B. $\vth^\star$ is reached by $300$ AdamW steps (lr $2\cdot10^{-4}$, PyTorch defaults otherwise), batch size $8$, with each batch item's task drawn uniformly from the five tasks. Everything runs in float32 with eager attention, deterministic algorithms enabled, and seeds $\{0,1,2\}$ controlling both PyTorch and NumPy RNG streams. In the full-fine-tuning control all parameters are trainable (no adapter) at lr $2\cdot10^{-5}$.

\paragraph{Tasks and prompts.}
Training pools are the first $2{,}000$ train examples per task; evaluation pools the first $128$ validation examples. Sequences are truncated to $160$ tokens; the loss is the HuggingFace token-mean cross-entropy over continuation tokens only (prompt tokens masked). Templates (\texttt{\textbackslash n} = newline; label continuations shown after the arrow, each with a leading space):
\begin{compactitem}
\item SST-2: \texttt{Review: \{sentence\}\textbackslash nSentiment:} $\to$ \texttt{negative}~/~\texttt{positive};
\item BoolQ: \texttt{\{passage[:600]\}\textbackslash nQuestion: \{question\}?\textbackslash nAnswer:} $\to$ \texttt{no}~/~\texttt{yes};
\item RTE: \texttt{\{premise\}\textbackslash nQuestion:}\allowbreak\ \texttt{Does this entail: "\{hypothesis\}"?\textbackslash nAnswer:} $\to$ \texttt{yes}~/~\texttt{no};
\item ARC-Easy: \texttt{Question: \{question\}\textbackslash nAnswer:} $\to$ each answer choice as a continuation;
\item HellaSwag: the context as prompt $\to$ the four endings as continuations.
\end{compactitem}

\paragraph{The probe set.}
$L$, $\vg$, and $H$ are all computed on the same four frozen multitask batches of eight examples ($32$ examples total), sampled task-uniformly from the training pools by a dedicated RNG stream fixed per (model, seed). $L$ is the mean of the four batch token-mean losses; gradients and Hessian-vector products average the per-batch quantities with equal batch weights (token-weighted within batch, which reproduces each batch's token-mean exactly). An earlier draft described this probe as ``$128$ examples,'' conflating it with the per-task evaluation-pool size; the probe is $32$ examples.

\paragraph{Per-property settings.}
\prop{P1}: $\hat\vg$ is the unit probe gradient; $L(\vth^\star\pm\sigma\hat\vg)$ is evaluated on the probe set at each grid $\sigma$. \prop{P2}: $m=32$ per-minibatch gradients (batch size $8$, fresh task-uniform draws), centered; $r_{90}$ from the singular values, so the rank ceiling is $m-1$; the $m$-sweep control repeats this at $m\in\{32,64,128,256\}$ (seed $0$) with an isotropic null estimated from Gaussian matrices of matched $m$ (three repetitions). \prop{P3}: BoolQ trajectory from $\vth^\star$ under AdamW at lr $2\cdot10^{-4}$; at each checkpoint the top-$10$ subspace of the single-task gradient covariance is estimated twice from $32$ gradients each; the resampling floor is the overlap of the two same-checkpoint estimates, and the onset is the first checkpoint where the overlap with the step-8 subspace falls below the floor minus $0.02$. \prop{P4}: task vectors are $40$-step AdamW single-task fine-tunes from $\vth^\star$; $\vh(\vdel)$ is the mean over $16$ fixed SST-2 training prompts of the last-token hidden state at the output of block $\lfloor0.55\,n_L\rfloor$ (e.g.\ block $13$ of $24$ for GPT-Neo-1.3B); the null applies random weight deltas matched in norm to each task vector. \prop{P5}: the task is SST-2; the candidate layers are the outputs of blocks $\lfloor0.4\,n_L\rfloor$, $\lfloor0.55\,n_L\rfloor$, and $\lfloor0.7\,n_L\rfloor$ (blocks $9/13/16$ for a $24$-block model), the ``upper-medium'' set; the CAA vector is built from $24$ validation prompts as the mean prompt-only last-token hidden state of positive-label minus negative-label examples, per example without padding; the weight move is one gradient step $-\eta \vg_{\rm sst2}$ at $\eta=10^{-4}$ from a single-minibatch SST-2 gradient; the activation shift is the mean per-prompt last-token change; the reported cosine is the best over the three layers, and the null repeats the measurement with a norm-matched random weight perturbation. \prop{P6}: $256$ draws $\vdel\sim\mathcal N(0,\sigma^2I)$ at per-coordinate $\sigma=10^{-3}$; gains are probe-set loss differences; the prediction is $-\vg^\top\vdel$ with $\vg$ the probe gradient. \prop{P7}: $\vv$ is the unit probe gradient and the comparator one random unit vector; curvatures are $\vv^\top H\vv$ via double-backward HVPs on the probe set; the reconstruction check compares $\vv^\top H\vv$ to the gradient finite difference $v^\top(\vg(\vth^\star+\epsilon\vv)-\vg(\vth^\star-\epsilon\vv))/2\epsilon$ at $\epsilon\in\{10^{-5},10^{-6},10^{-4}\}$, accepting the first $\epsilon$ with ratio in $[0.5,2]$. \prop{P8}: each one-step update uses a fresh single-minibatch task gradient from a fixed per-task RNG stream, the same stream for both orderings, so the two paths differ only in order; the coarse onset is the smallest grid $\eta$ with $c(\eta)\ge0.10$.

\paragraph{Commutator measurements.}
Task gradients and HVPs in Section~\ref{sec:spine} use three frozen single-task batches per task (a fixed per-task RNG stream), shared between the two-step defect and the $\kappa$ commutator: the ``matched estimator.'' The top-$10$ Hessian subspace and $\lambda_{\max}$ come from Lanczos with $24$ iterations and full reorthogonalization on the probe-set HVP operator. The fine grid is $25$ logarithmically spaced points on $[10^{-5},10^{-1}]$ ($33$ points from $10^{-7}$ for the full-fine-tuning control, whose $\kappa$ is larger), with $\eta^\dagger$ read by log-log interpolation of the $c=0.10$ crossing and left/right censoring recorded per cell. Parameters are restored to $\vth^\star$ between pairs; an early run that failed to do so was caught by the pre-registered checks and fully re-run. The estimator-decomposition run replaces only the defect-side gradients with fresh single-minibatch estimators, leaving the $\kappa$ side at three batches.

\paragraph{Cross-fitted functional confirmation.}
The confirmation models are GPT-2-large, OLMo-2-7B, Mistral-7B-v0.3, and Qwen2.5-14B, with seeds $\{0,1,2\}$. The first $2{,}000$ valid training examples per task define the common operating point; valid training ranks $2{,}129$--$2{,}188$ define geometry and ranks $2{,}189$--$2{,}248$ define the independent update paths. Evaluation uses valid validation ranks $257$--$384$, except RTE uses disjoint training ranks $2{,}249$--$2{,}376$ because its validation split is shorter. Every task therefore has $60$ geometry examples, $60$ update examples, and $128$ evaluation examples. The ten task pairs and five evaluation tasks are scored at target effect $0.05$ with $\eta=0.05/\kappa_G$ and admissible interval $[10^{-5},3\cdot10^{-2}]$; cells outside it are emitted at the endpoint but excluded from primary aggregates. The primary rules require overall $R^2_{\rm zero}>0$ with at least three of four models positive, overall endpoint cosine $>0.50$ with at least three models positive, superiority to a fixed cyclic pair shuffle on both quantities, and at most $25\%$ step-clipped pair cells. All choices, floating-point tolerances, and the rule that the raw-output check completes before any scoring were frozen before confirmation contact.

\paragraph{Extended-sweep run.}
Per task at $\vth^\star$: a fixed per-task probe set of four eight-example training batches; the per-task gradient is the four-batch average; the loss sweep covers $13$ logarithmically spaced signed scales $s\in\pm[10^{-4},1]$ along the unit task gradient and along two fixed random unit directions shared across tasks; multiple-choice accuracy (argmin of per-choice continuation token-mean loss, first $55$--$64$ usable validation examples) is probed at $\vth^\star$ and at $\vth^\star\pm s\hat\vg_t$ for $s\in\{10^{-2},10^{-1},0.3,1\}$; parameters are restored to $\vth^\star$ after every sweep point.

\paragraph{Follow-up controls.}
The \prop{P7} distribution uses $64$ random unit directions per cell (a fresh generator per cell) with one probe HVP each, recording signed $\vr^\top H\vr$ and $\norm{H\vr}$. The \prop{P6} sweep repeats the $256$-draw protocol at each of the seven grid scales, with a fresh generator per scale. The effective-weight control materializes each task's $\Delta W$ from the LoRA factors per adapted module, applies $\Delta W_A+\Delta W_B$ to the base weights with the adapters held at $\vth^\star$, and requires the two routes to agree on single tasks to relative error $10^{-2}$ (all $81$ cells pass). The accuracy rerun repeats the two archived best-of-$256$ search cells with frozen radii and search seeds, and records multiple-choice accuracy (the extended-sweep rule) on the first $128$ loaded validation examples per task, at $\vth^\star$ and at each selected perturbation.

\paragraph{Compute.}
All sweeps ran on A100 80GB GPUs, one (model, seed) cell per GPU, in float32 with eager attention and offline HuggingFace caches. At $13$--$14$B the gradient/HVP closures are row-chunked (microbatch $2$) with shifted-active-token weighting, which reproduces the batch token-mean exactly, so chunking changes peak memory only. The best-of-$N$ search demo draws $256$ perturbations per radius multiplier ($\{0.25,0.5,1,2,4\}\times r_{\rm forecast}$, where $r_{\rm forecast}=\eta_{\rm forecast}\norm{\vg}$) and evaluates target and side-task losses on fixed probe batches.

\section{Showcase accuracy per cell}
\label{app:showcaseacc}
Table~\ref{tab:showcaseacc} tabulates multiple-choice accuracy for all $15$ (model, task) cells of the extended sweeps of Section~\ref{sec:firstlook}: the base value at $\vth^\star$ and the signed change at every probed scale, in both directions along the task gradient. It is a descriptive tabulation of the archived extended-sweep outputs. With $55$--$64$ usable examples per task, one example is worth $1.6$--$1.8$pp; entries of that size are counting noise. Responses above $10$pp appear only at $|s|\ge0.1$ and only on five of the $15$ cells (both BoolQ cells, SST-2 on Pythia-410M and on Qwen2.5-7B, and ARC-Easy on Qwen2.5-7B at $s=1$); the other ten cells stay within a few examples of their base value even at $s=1$.

\begin{table}[!t]
\centering\footnotesize
\setlength{\tabcolsep}{4.5pt}
\caption{Multiple-choice accuracy across the extended-sweep grid: base accuracy at $\vth^\star$ (percent) and change (pp) at each probed scale $s$ along $\mp\hat\vg_t$; $n$ is the number of usable evaluation examples.}
\label{tab:showcaseacc}
\begin{tabular}{llrr rrrr rrrr}
\toprule
 & & & & \multicolumn{4}{c}{downhill $-s$} & \multicolumn{4}{c}{uphill $+s$} \\
\cmidrule(lr){5-8}\cmidrule(lr){9-12}
model & task & $n$ & base \% & $1.0$ & $0.3$ & $0.1$ & $0.01$ & $0.01$ & $0.1$ & $0.3$ & $1.0$ \\
\midrule
neo-1.3B & \task{sst2} & $64$ & $90.6$ & $-1.6$ & $-1.6$ & $-3.1$ & $+0.0$ & $+1.6$ & $+1.6$ & $+0.0$ & $-4.7$ \\
neo-1.3B & \task{boolq} & $59$ & $72.9$ & $+1.7$ & $+1.7$ & $+1.7$ & $+0.0$ & $+1.7$ & $-5.1$ & $-35.6$ & $-47.5$ \\
neo-1.3B & \task{rte} & $61$ & $59.0$ & $-9.8$ & $-9.8$ & $-1.6$ & $+3.3$ & $-3.3$ & $-8.2$ & $-8.2$ & $-8.2$ \\
neo-1.3B & \task{arc-e} & $64$ & $64.1$ & $+3.1$ & $-1.6$ & $+0.0$ & $+0.0$ & $+0.0$ & $-1.6$ & $-3.1$ & $-3.1$ \\
neo-1.3B & \task{hella} & $64$ & $43.8$ & $-4.7$ & $-4.7$ & $+0.0$ & $+0.0$ & $+0.0$ & $+0.0$ & $-3.1$ & $-3.1$ \\
pythia-410m & \task{sst2} & $64$ & $87.5$ & $-39.1$ & $-21.9$ & $-3.1$ & $+0.0$ & $+0.0$ & $-14.1$ & $-35.9$ & $-35.9$ \\
pythia-410m & \task{boolq} & $59$ & $47.5$ & $+25.4$ & $+25.4$ & $+25.4$ & $+10.2$ & $-6.8$ & $-20.3$ & $-20.3$ & $-20.3$ \\
pythia-410m & \task{rte} & $61$ & $59.0$ & $-8.2$ & $-6.6$ & $+0.0$ & $+1.6$ & $-1.6$ & $-4.9$ & $-9.8$ & $-9.8$ \\
pythia-410m & \task{arc-e} & $64$ & $50.0$ & $+0.0$ & $+4.7$ & $+0.0$ & $+0.0$ & $+0.0$ & $-1.6$ & $-1.6$ & $+3.1$ \\
pythia-410m & \task{hella} & $64$ & $37.5$ & $-6.2$ & $+0.0$ & $+0.0$ & $-1.6$ & $+0.0$ & $+1.6$ & $+0.0$ & $+1.6$ \\
Qwen2.5-7B & \task{sst2} & $64$ & $96.9$ & $-18.8$ & $+1.6$ & $+3.1$ & $+1.6$ & $+0.0$ & $+0.0$ & $-1.6$ & $-15.6$ \\
Qwen2.5-7B & \task{boolq} & $55$ & $89.1$ & $+1.8$ & $-1.8$ & $+1.8$ & $+0.0$ & $+0.0$ & $+0.0$ & $+0.0$ & $-3.6$ \\
Qwen2.5-7B & \task{rte} & $60$ & $90.0$ & $-1.7$ & $+0.0$ & $+0.0$ & $+0.0$ & $+0.0$ & $+0.0$ & $+1.7$ & $-5.0$ \\
Qwen2.5-7B & \task{arc-e} & $64$ & $87.5$ & $+0.0$ & $+1.6$ & $+0.0$ & $+0.0$ & $+0.0$ & $+0.0$ & $-1.6$ & $-15.6$ \\
Qwen2.5-7B & \task{hella} & $64$ & $67.2$ & $-6.2$ & $-3.1$ & $+0.0$ & $+0.0$ & $+0.0$ & $+0.0$ & $+0.0$ & $-3.1$ \\
\bottomrule
\end{tabular}

\end{table}

\section{Threshold sensitivity}
\label{app:thresholds}
The thresholds are selection bars; changing any bar after seeing the data would void the selection protocol. As a sensitivity check, Table~\ref{tab:thresholds} reruns the stored development and held-out summaries under nearby bars without reselecting any property. The headline pattern is stable: \prop{P1} and \prop{P6} keep under all tested bars; \prop{P5} drops under all tested bars; \prop{P4} remains scale-limited because relaxing the additivity bar can improve the development count but still leaves only $1/3$ held-out models passing at full task-vector scale. The \prop{P7} rows are magnitude-only sensitivity checks and do not apply the instrument gate, so they do not change the clean-evidence count.

\begin{table}[!t]
\centering\footnotesize
\caption{Threshold sensitivity on the stored atlas measurements: the original development-plus-held-out rule applied to nearby bars, one diagnostic at a time.}
\label{tab:thresholds}
\begin{tabular}{llccc}
\toprule
property & threshold & dev pass & holdout pass & rule verdict \\
\midrule
\prop{P1} & 0.05 & 5/6 & 3/3 & keep \\
\prop{P1} & 0.1 & 6/6 & 3/3 & keep \\
\prop{P1} & 0.2 & 6/6 & 3/3 & keep \\
\prop{P4} & 0.1 & 2/6 & 0/3 & drop \\
\prop{P4} & 0.15 & 4/6 & 1/3 & drop \\
\prop{P4} & 0.2 & 5/6 & 1/3 & drop \\
\prop{P5} & 0.2 & 1/6 & 0/3 & drop \\
\prop{P5} & 0.3 & 0/6 & 0/3 & drop \\
\prop{P5} & 0.4 & 0/6 & 0/3 & drop \\
\prop{P6} & 0.3 & 6/6 & 3/3 & keep \\
\prop{P6} & 0.5 & 6/6 & 3/3 & keep \\
\prop{P6} & 0.7 & 5/6 & 3/3 & keep \\
\prop{P7} magnitude & 2 & 6/6 & 3/3 & keep \\
\prop{P7} magnitude & 3 & 6/6 & 3/3 & keep \\
\prop{P7} magnitude & 10 & 6/6 & 3/3 & keep \\
\bottomrule
\end{tabular}

\end{table}

\section{Per-seed values of the headline statistics}
\label{app:seeds}
Per-seed values (seeds $0/1/2$) behind every per-model median in Table~\ref{tab:atlas}; `---' marks a value the harness did not produce for that seed. Radii are grid-quantized, so identical seed values are expected where the onset falls in the same grid cell.
\begin{table}[h]\centering\tiny\setlength{\tabcolsep}{1.8pt}
\caption{Per-seed values, all nine models (development then held-out).}
\resizebox{\textwidth}{!}{%
\begin{tabular}{lcccccccc}
\toprule
model & \prop{P1} $\sigma^\star_1$ & \prop{P2} $r_{90}$ & \prop{P3} onset & \prop{P4} $\varepsilon(1)$ & \prop{P5} $\cos$ & \prop{P6} corr & \prop{P7} $|$ratio$|$ & \prop{P8} $\eta^\dagger$ \\
\midrule
distilgpt2 & 1e-2/1e-2/1e-2 & 23/22/23 & 128/---/32 & 0.10/0.06/0.07 & +0.13/+0.01/-0.09 & 0.77/0.75/0.80 & 1.5e+03/3.5e+03/1.7e+03 & 1e-5/1e-5/1e-5 \\
pythia-160m & 1e-4/1e-2/1e-2 & 23/24/24 & 16/16/32 & 0.10/0.14/0.18 & +0.01/+0.13/+0.01 & 0.81/0.74/0.70 & 1.7e+05/3.4e+05/6.8e+02 & 3e-4/3e-5/3e-5 \\
pythia-410m & 1e-2/1e-2/1e-2 & 24/24/25 & 16/16/16 & 0.10/0.08/0.06 & -0.01/-0.12/-0.16 & 0.97/0.97/0.97 & 3.1e+04/8.2e+04/2.9e+03 & 3e-3/3e-3/3e-3 \\
neo-1.3B & 1e-2/1e-2/1e-2 & 20/21/23 & 16/32/32 & 0.19/0.21/0.19 & +0.05/+0.34/-0.04 & 1.00/1.00/1.00 & 1.0e+04/1.4e+05/1.9e+04 & ---/1e-2/--- \\
opt-1.3b & 1e-2/1e-2/1e-2 & 23/21/23 & 16/16/32 & 0.35/0.40/0.25 & +0.34/-0.15/+0.22 & 0.84/0.70/0.67 & 4.6e+04/9.6e+04/7.0e+04 & 1e-5/1e-5/1e-5 \\
TinyLlama-1.1B & 1e-2/1e-2/1e-2 & 23/21/24 & 16/16/32 & 0.14/0.14/0.15 & +0.04/+0.04/+0.07 & 1.00/1.00/1.00 & 4.2e+05/2.1e+05/1.5e+04 & 3e-3/1e-2/1e-2 \\
pythia-1.4B & 1e-2/1e-2/1e-2 & 23/23/24 & 16/16/16 & 0.11/0.08/0.22 & +0.01/-0.23/-0.22 & 1.00/1.00/1.00 & 6.2e+03/4.9e+04/1.7e+04 & 1e-2/3e-3/1e-2 \\
Qwen2.5-7B & 1e-2/1e-2/1e-2 & 23/23/25 & 16/32/16 & 0.36/0.26/0.32 & +0.18/-0.05/+0.01 & 1.00/1.00/1.00 & 6.4e+04/4.0e+04/2.1e+02 & 1e-2/1e-2/--- \\
OLMo-7B & 1e-2/1e-2/1e-2 & 25/11/21 & 16/16/16 & 0.31/0.38/0.32 & +0.01/-0.09/-0.06 & 0.99/0.97/0.97 & 1.8e+05/1.2e+07/2.4e+05 & 1e-2/3e-3/3e-3 \\
\bottomrule
\end{tabular}}
\end{table}

\end{document}